\documentclass[iicol,pdflatex,sn-mathphys-num]{sn-jnl}
\usepackage{array}          
\usepackage{multirow}       
\usepackage{hhline}         
\usepackage{booktabs}       

\usepackage{amsmath,amssymb,amsfonts} 
\usepackage{amsthm}
\usepackage{mathrsfs}
\usepackage[title]{appendix}
\usepackage{textcomp}
\usepackage{manyfoot}
\usepackage{algorithm}
\usepackage{algorithmicx}
\usepackage{algpseudocode}
\usepackage{listings}
\usepackage{xspace}

\usepackage{graphicx}
\usepackage[export]{adjustbox}
\usepackage{makecell}
\usepackage{url}
\usepackage{hyperref}
\usepackage{lineno}
\usepackage{caption}

\usepackage{tikz}
\usepackage[misc,geometry]{ifsym}
\usepackage{mathtools}

\usepackage[table]{xcolor} 

\definecolor{bestColor}{RGB}{255, 0, 0}    
\definecolor{secondBestColor}{RGB}{0, 0, 255} 
\definecolor{thirdBestColor}{RGB}{240,0, 240}  
\definecolor{qBestColor}{RGB}{0,215, 0}  

\definecolor{secondBestColor}{RGB}{0, 0, 255} 
\definecolor{thirdBestColor}{RGB}{240,0, 240}  
\definecolor{qBestColor}{RGB}{0,215, 0}  

\definecolor{cnn_color}{RGB}{240,248,255}  
\definecolor{incompression_color}{RGB}{255,250,240}  
\definecolor{crosscompression_color}{RGB}{240,255,240}  
\definecolor{input_color}{RGB}{255,240,240}      
\definecolor{mamba_color}{RGB}{235,235,225}  

\definecolor{deepred}{RGB}{133,0,0}    
\definecolor{lightred}{RGB}{255,102,102}  
\definecolor{deepgreen}{RGB}{0,133,0}  

\definecolor{bestcrossmanupulation_color}{RGB}{255,240,240} 
\definecolor{secondcrossmanupulation_color}{RGB}{240,240,255} 
\definecolor{thirdcrossmanupulation_color}{RGB}{240,255,240} 
\definecolor{background_color}{RGB}{255,255,255}

\newcommand{\addblankpages}[1]{
    \ifnum\value{page}>1\newpage\fi
    \begingroup
    \pagestyle{empty}
    \count0=#1
    \loop
        \ifnum\count0>0
            \null
            \newpage
            \advance\count0 by -1
    \repeat
    \endgroup
}

\theoremstyle{thmstyleone}%
\theoremstyle{thmstyletwo}%

\theoremstyle{thmstylethree}%

\begin{document}

\title{UMCL: Unimodal-generated Multimodal Contrastive Learning for Cross-compression-rate Deepfake Detection}

\author[1]{\fnm{Ching-Yi} \sur{Lai}}\email{ching1999.work@gmail.com}

\author[2]{\fnm{Chih-Yu} \sur{Jian}}\email{ru0354m3@gmail.com}

\author[3]{\fnm{Pei-Cheng} \sur{Chuang}}\email{sandy.purin@gmail.com}

\author[2]{\fnm{Chia-Ming} \sur{Lee}}\email{zuw408421476@gmail.com}

\author[2,4]{\fnm{Chih-Chung} \sur{Hsu}}\email{chihchung@nycu.edu.tw}

\author[1]{\fnm{Chiou-Ting} \sur{Hsu}}\email{cthsu@cs.nthu.edu.tw}

\author*[3]{\fnm{Chia-Wen} \sur{Lin}}\email{cwlin@ee.nthu.edu.tw}

\affil[1]{\orgdiv{Dept. Computer Science}, \orgname{National Tsing Hua University}, \orgaddress{ \city{Hsinchu} \postcode{300044}, \country{Taiwan}}}

\affil[2]{\orgdiv{Institute of Data Science}, \orgname{National Cheng Kung University}, \orgaddress{ \city{Tainan} \postcode{701401}, \country{Taiwan}}}

\affil[3]{\orgdiv{Department of Electrical Engineering}, \orgname{National Tsing Hua University},\orgaddress{ \city{Hsinchu} \postcode{300044}, \country{Taiwan}}}

\affil[4]{\orgdiv{Institute of Intelligent Systems}, \orgname{National Yang Ming Chiao Tung University}, \orgaddress{ \city{Tainan} \postcode{711010}, \country{Taiwan}}}

\abstract{

In deepfake detection, the varying degrees of compression employed by social media platforms pose significant challenges for model generalization and reliability. Although existing methods have progressed from single-modal to multimodal approaches, they face critical limitations: single-modal methods struggle with feature degradation under data compression in social media streaming, while multimodal approaches require expensive data collection and labeling and suffer from inconsistent modal quality or accessibility in real-world scenarios.
To address these challenges, we propose a novel Unimodal-generated Multimodal Contrastive Learning (UMCL) framework for robust cross-compression-rate (CCR) deepfake detection. In the training stage, our approach transforms a single visual modality into three complementary features: compression-robust rPPG signals, temporal landmark dynamics, and semantic embeddings from pre-trained vision-language models. These features are explicitly aligned through an affinity-driven semantic alignment (ASA) strategy, which models inter-modal relationships through affinity matrices and optimizes their consistency through contrastive learning. Subsequently, our cross-quality similarity learning (CQSL) strategy enhances feature robustness across compression rates.
Extensive experiments demonstrate that our method achieves superior performance across various compression rates and manipulation types, establishing a new benchmark for robust deepfake detection. Notably, our approach maintains high detection accuracy even when individual features degrade, while providing interpretable insights into feature relationships through explicit alignment.
}


\maketitle

	
    \section{Introduction}
	\label{sec:intro}
	
In an era where deepfake technologies can seamlessly fabricate digital media, maintaining trust in online content has become a critical societal challenge. Advanced generative models, such as GANs~\cite{goodfellow2014generativeadversarialnetworks} and diffusion models~\cite{ho2020denoisingdiffusionprobabilisticmodels}, enable manipulators to create highly realistic videos that evade traditional detection methods. While numerous single-modality detection approaches have been proposed~\cite{uia-vit,qian2020thinking,DFIL,hernandezortega2020deepfakesonphysdeepfakesdetectionbased,li2018ictu,xu2024tallthumbnaillayoutdeepfake}, they face a critical limitation in real-world deployment scenarios.

\begin{figure*}[h]
   \centering
   \includegraphics[width=1.0\textwidth]{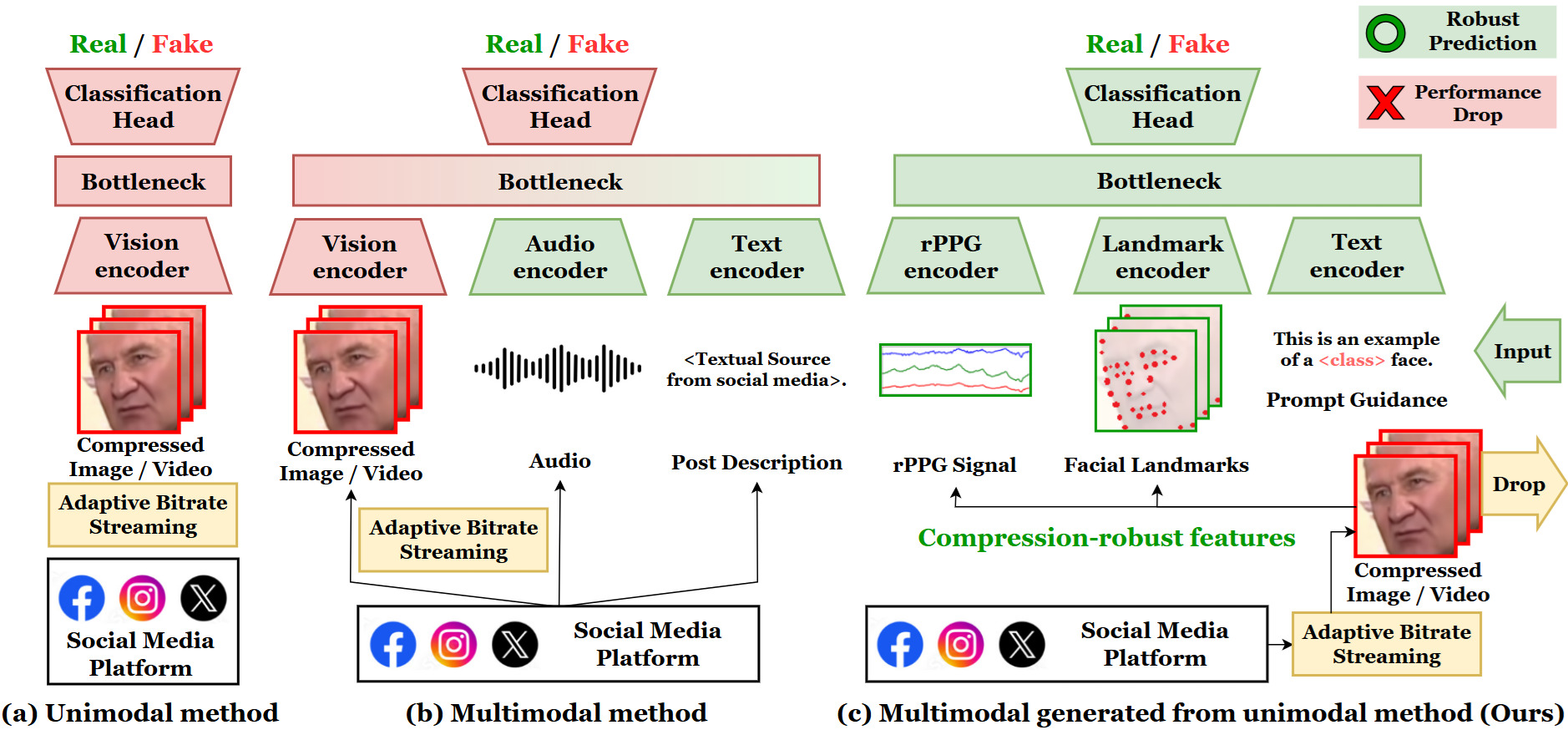}
 \caption{Paradigm comparison between deepfake detection methods. (a) Unimodal methods rely on visual features susceptible to compression artifacts. (b) Traditional multimodal methods require costly data collection and suffer from unequal modality degradation. (c) Our UMCL derives compression-robust multimodal features (rPPG, landmarks, text embeddings) from a single visual source, ensuring consistent modality availability across compression scenarios.}
   \label{fig:paradigm}
\end{figure*}

Multimedia content inevitably undergoes quality degradation during capture, transmission, and distribution, particularly on social media platforms where compression levels vary widely. While these compression-induced distortions are often imperceptible to human observers, they can significantly degrade model performance, posing a cross-compression-rate (CCR) challenge: detection models trained on high-quality data often fail to generalize to the heavily compressed videos encountered in real-world scenarios~\cite{5688237,shao2024deepfakeadapter}. State-of-the-art models frequently suffer severe performance drops when applied to such compressed content. The problem is further complicated by the heterogeneous compression strategies employed by different platforms. Services like YouTube and Facebook~\cite{facebook_video_encoding} use distinct transcoding pipelines, resulting in unpredictable and diverse quality distributions~\cite{motivation1}. This platform-induced variability introduces an "unknown condition", where the exact compression history of a video is unavailable, making it impossible to annotate training data accordingly. As a result, a fundamental domain gap arises between training and inference conditions, severely limiting practical deployment.

Recent research has explored different solutions to this challenge. As illustrated in Fig.~\ref{fig:paradigm}, single-modality approaches~\cite{QAD,DFL} attempt to learn compression-invariant features, while traditional multimodal methods integrate complementary features from multiple data sources. However, single-modality methods remain vulnerable to compression artifacts, and multimodal approaches face practical deployment challenges: costly data collection~\cite{shao2023dgm4}, unequal modality degradation, and potential modality unavailability in real-world scenarios. Our prior work CPML~\cite{cpml} pioneered generating multiple modalities from a single visual source, demonstrating the potential of this paradigm.

To directly tackle the CCR challenge, we extend CPML~\cite{cpml} and propose the Unimodal-generated Multimodal Contrastive Learning (UMCL) framework. UMCL is designed to extract compression-resilient multimodal features from a single visual modality by transforming input videos into three complementary representations: physiological signals (rPPG), facial landmark dynamics, and semantic embeddings derived from text prompts. This design offers several key advantages: eliminating costly multimodal data collection, ensuring modality synchronization, maintaining consistent data availability, and effectively mitigating the domain gaps that typically cause performance degradation in traditional models.

Although UMCL builds upon CPML~\cite{cpml}, our analysis reveals critical vulnerabilities in CPML’s attention-based fusion mechanism. Specifically, CPML often exhibits uneven feature dependencies across modalities, leading to semantic inconsistencies and over-reliance on dominant input sources. As a result, its performance degrades significantly when feature quality deteriorates, particularly under varying compression conditions.

To overcome these limitations, UMCL introduces two novel strategies. First, the Affinity-driven Semantic Alignment (ASA) explicitly models inter-modal relationships via affinity matrices, enabling compact and semantically coherent representations that remain stable even when individual modalities are degraded. Second, the Cross-Quality Similarity Learning (CQSL) enforces feature consistency across different compression levels, thereby enabling joint learning across both modality and quality domains. Together, ASA and CQSL enable UMCL to construct unified, robust representations that markedly improve resilience to modality degradation and compression variability.

UMCL generates complementary modalities with distinct compression robustness: rPPG signals capture stable temporal patterns, facial landmarks preserve structural geometry, and semantic embeddings provide high-level contextual features that remain resilient to low-level visual artifacts. Leveraging the synergistic operation of ASA and CQSL, UMCL effectively enforces both complementarity and consistency among these self-generated modalities. This unified framework enables robust representation learning without relying on external data sources, thereby overcoming the deployment challenges that limit traditional multimodal approaches under diverse compression conditions.

\indent
Our key contributions are as follows:
\begin{itemize}
   \item We propose the UMCL framework, which derives three complementary modalities---rPPG signals, facial landmark dynamics, and semantic text prompts---from a single visual input, eliminating the reliance on multiple data sources for CCR deepfake detection while ensuring inherent modality synchronization and consistent data availability.
   \item We introduce the ASA strategy, which addresses the vulnerabilities of attention-based fusion mechanisms by explicitly modeling inter-modal relationships through affinity matrices. ASA aggregates multimodal features into a compact and coherent feature space, ensuring semantic consistency and preventing over-reliance on single modalities during feature degradation.
   \item We develop the CQSL strategy, which works synergistically with ASA to enhance feature consistency across different compression rates through contrastive learning on rPPG signals. This joint cross-modal and cross-compression learning maintains high detection performance despite variations in compression quality.
   \item Extensive experiments validate that UMCL significantly outperforms existing methods across various compression rates and deepfake manipulation types, establishing a new benchmark for robust deepfake detection. Notably, our approach maintains high detection accuracy even under severe modality degradation and demonstrates superior robustness compared to our preliminary CPML version, while providing interpretable insights into cross-modal feature interactions.
\end{itemize}

\indent
Building upon our preliminary CPML work~\cite{cpml}, this extended UMCL introduces a novel ASA strategy to overcome CPML’s vulnerabilities in attention-based fusion and integrates CQSL with additional physiological constraints to ensure reliable heart rate estimation. With the new ASA alignment, UMCL achieves clean-condition performance comparable to CPML while delivering substantial robustness gains under realistic modality degradation. Comprehensive experiments and ablation studies confirm that these enhancements significantly improve resilience to modality degradation, resulting in more stable and generalizable deepfake detection across diverse real-world scenarios. By effectively addressing the CCR challenge through unimodal-generated multimodal learning, UMCL establishes a new state of the art in cross-compression-rate deepfake detection and provides a practical solution for deployment in social media environments characterized by heterogeneous compression levels and data inconsistencies.

\indent The rest of this paper is organized as follows. Most relevant works are surveyed in Sec.~\ref{sec:review}. Sec.~\ref{sec:motivation} explains the motivations behind this work. The proposed schemes for multimodal feature learning, alignment, and fusion are elaborated in Sec.~\ref{sec:method}. In Sec.~\ref{sec:experiments}, experimental results and analysis are demonstrated. Finally, conclusions are drawn in Sec.~\ref{sec:conclusion}.
	\section{Related Work}
	\label{sec:review}

\subsection{Unimodal Deepfake Detection Method}

Deepfake detection is typically formulated as a binary classification task, distinguishing authentic from manipulated facial content. Contemporary detection methods can be broadly categorized into three approaches: physical/physiological analysis, signal-level artifact detection, and data-driven learning based on large-scale benchmark datasets~\cite{cai2024av, roessler2019faceforensics++, li2020celeb, dufour2019deepfakes, 10.1609/aaai.v38i5.28310}.

Physical and physiological analysis-based methods exploit inconsistencies in biological signals, offering robustness under challenging conditions due to their interpretable nature. Notable techniques include eye-blinking pattern analysis~\cite{li2018ictu}, rPPG signals~\cite{hernandezortega2020deepfakesonphysdeepfakesdetectionbased}, corneal specular highlights~\cite{hu2020exposing}, and facial dynamics~\cite{masi2020twobranch}. However, while these approaches are effective in many scenarios, they struggle against high-quality forgeries that can accurately mimic biological cues~\cite{haliassos2021lips}.

Artifact-based and data-driven methods detect signal-level inconsistencies using sophisticated computational modeling. Techniques include frequency-domain analysis~\cite{durall2020watch}, frequency and wavelet-based methods~\cite{qian2020thinking}, compression artifact detection~\cite{10197527}, convolutional neural networks (CNNs)~\cite{chollet2017xception}, and Vision Transformers (ViT)-based~\cite{uia-vit} data-driven methods. 

While these models achieve high accuracy on benchmark datasets, their performance often degrades significantly (by over 20\%) in cross-domain or compressed media scenarios~\cite{shao2024deepfakeadapter}, limiting their practical effectiveness in real-world applications. These challenges highlight the need for detection methods that can generalize across varying compression levels and data quality conditions, ensuring robust deepfake detection in diverse deployment environments.

\subsection{Multimodal Deepfake Detection Method}

Multimodal deepfake detection methods aim to overcome the limitations of single-modal approaches by integrating complementary signals from multiple modalities, such as visual, audio, and physiological data. Recent advances include audio-visual fusion frameworks, such as AVTENet~\cite{hashmi2023avtenet}, and modality-agnostic architectures~\cite{yu2023modalityagnostic}, which leverage audio-visual speech recognition (AVSR) discrepancies to enhance detection robustness. Adapter-based methods, like Deepfake-Adapter~\cite{shao2024deepfakeadapter}, introduce lightweight, modality-specific adapters into ViTs to improve domain adaptation under varying input conditions. Additionally, hierarchical reasoning models, such as HAMMER~\cite{shao2023dgm4}, exploit cross-modal inconsistencies through hierarchical alignment and reasoning strategies, further improving detection accuracy.

Despite their advantages, multimodal methods face significant challenges in real-world deployment. For instance, attention-based fusion models~\cite{wang2021m2tr} suffer accuracy drops exceeding 20\% when a modality (e.g., audio) undergoes severe compression degradation or is entirely absent~\cite{yang2024testtime}. Similarly, adapter-based models like Deepfake-Adapter, while improving robustness, still require consistent modality availability across training and testing scenarios. These limitations highlight the need for frameworks that retain the benefits of multimodal learning without explicit dependence on multiple data sources, motivating the development of our unimodal-generated multimodal approach.

\subsection{Feature Alignment in Multimodal Learning}

Feature alignment is fundamental to effective multimodal learning, ensuring the seamless integration of diverse modality-specific representations. Recent research in multimodal alignment has primarily explored three key directions: label-based alignment, structured alignment, and dynamic fusion strategies.

Label-based alignment methods, such as LAMM~\cite{wang2024lammprompt} and Multimodal Alignment Prompting (MmAP)~\cite{wang2024mmap}, enhance cross-modal semantic consistency using hierarchical loss functions and prompting-based strategies, achieving effective alignment under ideal conditions. Structured alignment frameworks, like Structure-CLIP~\cite{wang2024structureclip}, leverage scene-graph knowledge to establish semantically rich correspondences across modalities, improving representation quality.

Dynamic fusion methods, exemplified by CaffNet~\cite{lee2021looking}, explicitly model both global and local affinities between modalities (e.g., audio-visual streams), significantly improving robustness and interpretability. Additionally, contrastive learning approaches~\cite{liang2023contrastivelearning} mitigate modality gaps through explicit cross-modal regularization, reducing alignment inconsistencies.

Despite these advancements, existing feature alignment techniques largely assume stable and complete modality availability. This assumption breaks down in real-world scenarios, particularly in social media contexts where modalities frequently experience uneven degradation or complete absence. Label-based and structured alignment methods fail when a modality provides unreliable or incomplete signals, leading to misalignment and biased fusion. Similarly, dynamic fusion methods, such as CaffNet, while effective in cross-modal alignment, are not designed to address intra-modality alignment challenges that arise when complementary features originate from a single modality.

To overcome these limitations, we propose ASA, which explicitly models intra-modality correlations between complementary rPPG signal, facial landmark dynamics, and semantic embeddings---all derived from a single visual modality. Unlike existing methods restricted to inter-modal alignment, ASA ensures stable, interpretable, and robust feature fusion even under severe compression-induced degradation, effectively addressing the shortcomings of current alignment techniques.

	\section{Motivations}
	\label{sec:motivation}
	
\begin{figure}[h]
\centering
\includegraphics[width=0.48\textwidth]{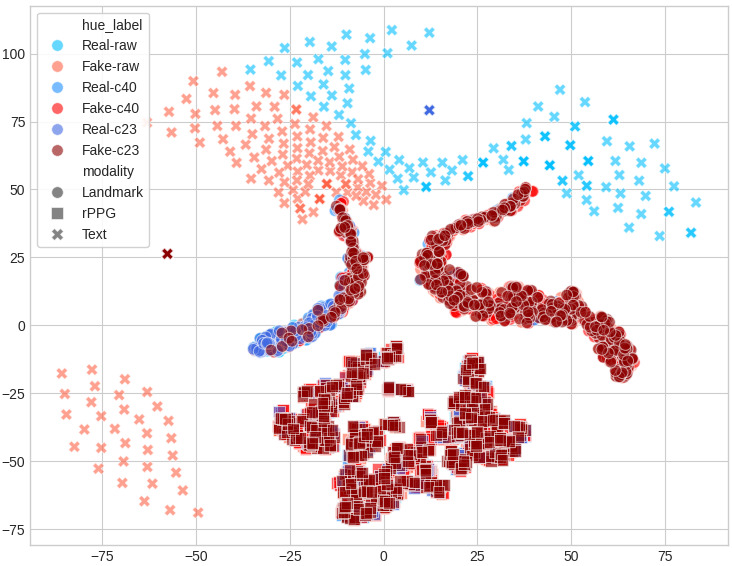}
\caption{Modality-wise feature visualization of CPML \cite{cpml} using t-SNE \cite{van2008visualizing}.}
\label{fig:tsne_motivation}
\end{figure}
While recent deepfake detection methods have demonstrated strong performance on high-quality datasets, they encounter critical challenges in real-world deployment. Multimedia content routinely undergoes quality degradation during distribution, particularly on social media platforms where compression levels vary widely. Although such distortions are often imperceptible to human observers, they substantially corrupt the visual features that detection models rely on, thereby creating significant domain gaps between training and inference. This problem is further complicated by heterogeneous compression strategies across platforms~\cite{facebook_video_encoding}, where distinct transcoding pipelines produce unpredictable quality distributions~\cite{motivation1}. Consequently, training–inference mismatches arising from these unknown compression conditions severely limit the practical deployment of existing detection models.
\\\indent
To address CCR challenge, our preliminary CPML method~\cite{cpml} pioneered a strategy of generating multiple compression-robust modalities---rPPG signals, facial landmarks, and text embeddings---from a single visual source. This approach circumvents both the compression vulnerability of traditional visual methods and the practical challenges of collecting synchronized multimodal data. However, our analysis reveals that CPML's attention-based fusion mechanism exhibits critical limitations in cross-modal coordination.
\\\indent
As illustrated in Fig.~\ref{fig:tsne_motivation}, the t-SNE visual of CPML's feature representations shows scattered and inconsistent distributions with poor separation between real and fake samples across different compression rates and manipulation types. This semantic inconsistency indicates insufficient inter-modal dependencies and coordination, potentially leading to over-reliance on certain modalities. When these relied-upon modalities degrade under compression, the model becomes vulnerable to robustness issues and performance drops.
\\\indent
These observations motivate our development of UMCL, which explicitly addresses the modal coordination problem through two novel strategies. The ASA strategy explicitly models inter-modal relationships to aggregate features into a more compact and coherent space, while the CQSL strategy ensures consistent representations across compression variations. 

	\section{Proposed Method}
	\label{sec:method}
	\subsection{Framework Overview}

\begin{figure*}[h]
	\centering
	\includegraphics[width=1.0\textwidth]{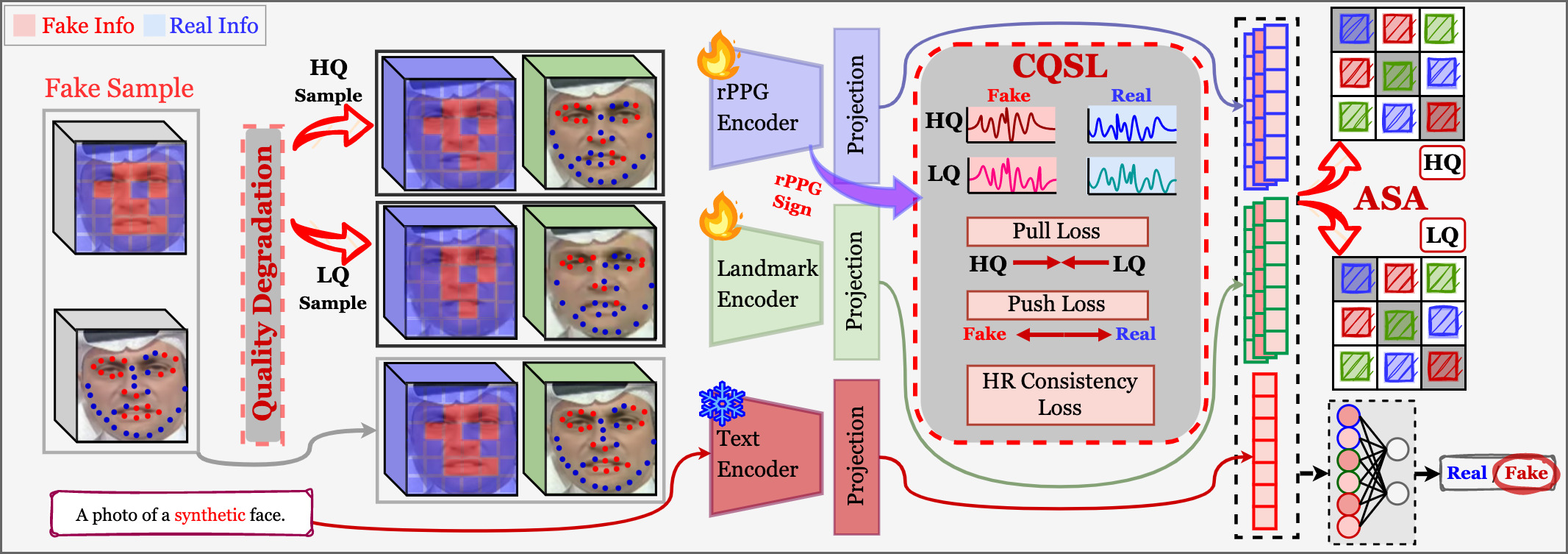}
  \caption{Overview of the proposed unimodal-generated multimodal contrastive learning (UMCL) framework for robust cross-compression rate deepfake detection.}
    \label{fig:framework}
    \vspace*{-4mm}
\end{figure*}


In this section we revisit the limitations of \emph{CPML}, whose cross–attention fusion is susceptible to compression‑specific dominance and lacks an explicit mechanism for aligning high‑ and low‑quality features.  
To overcome these issues, we propose the UMCL framework for cross‑compression‑rate (CCR) deepfake detection.  
As illustrated in Fig.~\ref{fig:framework}, UMCL derives three complementary modalities from a \emph{single} facial landmark trajectories, rPPG physiological signals, and semantic textual prompts—and processes them with a landmark encoder (L‑encoder), a physiological encoder (P‑encoder), and a text encoder (T‑encoder), respectively.  
The resulting embeddings are brought into a common semantic space by the proposed \emph{Affinity‑driven Semantic Alignment} (ASA) module, which eliminates the attention‑dominance bias observed in CPML, while \emph{Cross‑Quality Similarity Learning} (CQSL) explicitly ties high‑ and low‑quality feature pairs, closing the CCR gap. Symbols used throughout the paper are summarised in Table~\ref{table:notations}.

\begin{table}[t!]
\footnotesize
\caption{Notation}
\label{table:notations}
\renewcommand{\arraystretch}{1.2}  
\begin{tabular}{c|l}
\hline
\textbf{Symbol} & \textbf{Definition} \\
\hline
$\mathbf{X}^\mathrm{HQ/LQ}_\mathrm{class}$ & High/low-quality video sequence \\
$\mathbf{L}^\mathrm{HQ/LQ}_\mathrm{class}$ & High/low-quality facial landmark sequence \\
$\mathbf{z}^\mathrm{HQ/LQ}_\mathrm{class}$ & rPPG features from P-encoder \\
$\mathbf{e}^\mathrm{HQ/LQ}_\mathrm{class}$ & Landmark features from L-encoder \\
$\mathbf{t}^\mathrm{valid/invalid}_\mathrm{class}$ & Text embeddings from semantic prompts \\
$\hat{\mathbf{z}}, \hat{\mathbf{e}}, \hat{\mathbf{t}}$ & Projected feature vectors \\
$\Phi_p, \Phi_l, \Phi^\mathrm{frozen}_t$ & rPPG, landmark, and text encoders \\
$\mathbf{U}$ & Concatenated multimodal features \\
$\hat{\mathbf{Y}}$ & Classification output \\
${\mathbf{Y}}$ & Ground-truth \\
$\cal{L}_\mathrm{bce}$ & Classification loss \\
$\cal{L}_\mathrm{aff}$ & Semantic alignment loss \\
$\cal{L}^\mathrm{rPPG}_\mathrm{pull}$ & rPPG cross-quality discrepancy loss\\
$\cal{L}^\mathrm{rPPG}_\mathrm{push}$ & rPPG inter-class similarity loss \\
$\cal{L}_\mathrm{HR}$ & Heart rate consistency loss \\
$\cal{L}_\mathrm{phy}$ & Physiological loss functions \\
$\mathbf{A}^b$ & Inter-modality affinity matrix \\
$\hat{\mathbf{Z}}, \hat{\mathbf{E}}, \hat{\mathbf{T}}$ & Batch-level feature representations \\
$\mathbf{U}^b$ & Unified semantic embedding matrix \\
$T$ & The number of frames \\
$K$ & The length of rPPG signals \\
\hline
\end{tabular}
\end{table}

\subsection{Unimodal-generated Multimodal Contrastive Learning}
Traditional multimodal approaches, while effective, face significant challenges in data acquisition and modality alignment, especially under varying compression rates. To address these limitations while retaining the advantages of multimodal learning, we propose extracting complementary features from a single visual modality. This approach ensures inherent synchronization and robust accessibility across diverse compression scenarios. Our framework derives complementary features from three distinct sources: rPPG, temporal facial landmark dynamics, and semantic text prompts. Each modality contributes unique discriminative cues that, when properly aligned, enable resilient deepfake detection across different compression levels.

The P-encoder is designed to capture subtle blood flow patterns from facial regions—features that are often disrupted in synthetic videos generated by face forgery schemes. To extract the rPPG feature $\mathbf{z}$ from input videos, we employ the pre-trained PhysFormer~\cite{yu2022physformer}. Given a $T$-frame high-quality (HQ) video sequence $\mathbf{X}^{\mathrm{HQ}}_{\mathrm{class}} = \{\mathbf{x}^{i}_{\mathrm{class}}\}$, where $i=1, ..., T$,  $\mathrm{class} \in \{\mathrm{real}, \mathrm{fake}\}$, and $T=100$, we generate its low-quality (LQ) counterpart $\mathbf{X}^{\mathrm{LQ}}_{\mathrm{class}}$ by downsampling the HQ sequence by a factor of four in both spatial dimensions---simulating compression-induced degradation.

The P-encoder processes both HQ and LQ face regions of size $128 \times 128$, producing feature vectors $\mathbf{z}_{\mathrm{class}}^{\mathrm{HQ}}, \mathbf{z}_{\mathrm{class}}^{\mathrm{LQ}} \in \mathbb{R}^{1 \times 320}$ that encode temporal physiological patterns. To effectively capture multi-scale temporal variations, the model employs temporal strided convolutions followed by self-attention layers, allowing it to learn both rapid and slow changes in physiological signals ($\mathbf{z}^\mathrm{HQ}_\mathrm{real}$, $\mathbf{z}^\mathrm{LQ}_\mathrm{real}$, $\mathbf{z}^\mathrm{HQ}_\mathrm{fake}$, and $\mathbf{z}^\mathrm{LQ}_\mathrm{fake}$) across different compression rates:

\begin{equation}
\mathbf{z}_{\mathrm{class}}^{\mathrm{HQ/LQ}} = \Phi_{P}(\mathbf{X}^{\mathrm{HQ/LQ}}_{\mathrm{class}}), 
\label{eq:P-encoder}
\end{equation}
where $\Phi_{P}$ represents the P-encoder.

For facial landmark sequences, we employ a pre-trained LRNet~\cite{sun2021improving} as the L-encoder, which consists of two recurrent neural networks (RNNs) designed to model the temporal dynamics of facial landmarks, including their trajectories and motions, respectively.
Given a $T$-frame HQ landmark sequence $\mathbf{L}^{\mathrm{HQ}}_{\mathrm{class}} = \{l_{\mathrm{class}}^{i}\}$, where $i = 1, ..., T$, we introduce random positional shifts to generate its LQ counterpart:
\begin{equation} \mathbf{L}^{\mathrm{LQ}} = \mathbf{L}^{\mathrm{HQ}} + \Delta \mathbf{L}, \end{equation}
where $\Delta \mathbf{L}$ represents perturbations that simulate compression-induced degradation.

The L-encoder processes both HQ and LQ landmark sequences, extracting feature representations $\mathbf{e}_{\mathrm{class}}^{\mathrm{HQ}}, \mathbf{e}_{\mathrm{class}}^{\mathrm{LQ}} \in \mathbb{R}^{1 \times 128}$, which are subsequently fused to enhance feature robustness. This encoding mechanism preserves temporal coherence and improves generalization across varying compression rates:
\begin{equation}
\mathbf{e}_{\mathrm{class}}^{\mathrm{HQ/LQ}} = \Phi_{L}(\mathbf{L}^{\mathrm{HQ/LQ}}_{\mathrm{class}}), 
\label{eq:L-encoder}
\end{equation}
where $\Phi_{L}$ represents the L-encoder.

\begin{table*}[ht]\small
    \caption{Examples of textual prompts used for describing the real and manipulated categories, providing rich information for real/fake semantic understanding.}
    \label{table:prompts}
\scalebox{1.0}{
    \begin{tabular}{|c|c|} 
      \hline
      Real Prompts & Fake Prompts \\\hline
        'This is an example of a real face'  & 'This is an example of a forgery face' \\ 
        'This is a bonafide face' & 'This is an example of an attack face' \\
        'This is a real face' &'This is not a real face' \\
        'This is how a real face looks like' &'This is how a forgery face looks like' \\
        'a photo of a real face' &'a photo of a forgery face' \\
        'This is not a forgery face'  &'a printout shown to be a forgery face' \\\hline 
    \end{tabular}}
    \centering 
\end{table*}

Inspired by Visual-Language Pre-training~\cite{zhou2022conditional}, which has demonstrated strong generalization across varying image quality conditions, we incorporate language guidance using a pre-trained CLIP text encoder~\cite{radford2021learning}. As shown in Table~\ref{table:prompts}, we generate simple semantic text prompts for training face videos based on their real/fake labels. These text prompts are then processed by the T-encoder, which encodes semantic descriptions of real and manipulated facial characteristics, denoted as $\mathbf{T}_\mathrm{real}$ and $\mathbf{T}_\mathrm{fake}$, respectively.

By leveraging the CLIP text encoder, we map these textual descriptions into semantic embeddings that provide high-level contextual understanding. For each class, we maintain both original prompts $\mathbf{T}^{\mathrm{valid}}_\mathrm{\mathrm{class}}$, and randomly generated invalid prompts $\mathbf{T}^{\mathrm{invalid}}_\mathrm{\mathrm{class}}$  (e.g., random character sequences, irrelevant questions, or semantically misaligned descriptions) to enhance generalization and reduce overreliance on textual information. With equal probability (0.5), we use either the original semantic prompts or invalid ones. The T-encoder then generates semantic embeddings  $\mathbf{t}_{\mathrm{real}}^{\mathrm{valid}}, \mathbf{t}_{\mathrm{real}}^{\mathrm{invalid}}, \mathbf{t}_{\mathrm{fake}}^{\mathrm{valid}},
\mathbf{t}_{\mathrm{fake}}^{\mathrm{invalid}},  \in \mathbb{R}^{1 \times 512}$:
\begin{equation}
\mathbf{t}_{\mathrm{class}}^{\mathrm{valid/invalid}} = \Phi_{T}^{\mathrm{frozen}}(\mathbf{t}^{\mathrm{valid/invalid}}_{\mathrm{class}}),
\label{eq:T-encoder}
\end{equation}
where $\Phi_{T}^{\mathrm{frozen}}$ represents the pretrained,  parameter-frozen T-encoder.

After extracting modality-specific features at varying quality levels using specialized encoders, a key challenge is ensuring consistent feature distributions in a unified semantic space. To address this, we align the dimensionality and distribution of all features  {$\mathbf{z}_{\mathrm{class}}^{\mathrm{HQ/LQ}}$, $\mathbf{e}_{\mathrm{class}}^{\mathrm{HQ/LQ}}$, $\mathbf{t}_{\mathrm{class}}^{\mathrm{HQ/LQ}}$} using learnable projection layers. We adopt linear projection layers to balance computational efficiency and expressiveness, as validated in our experiments. This transformation yields projected feature representations $\hat{\mathbf{z}}_{{\mathrm{class}}}^{\mathrm{HQ/LQ}}, \hat{\mathbf{e}}_{{\mathrm{class}}}^{\mathrm{HQ/LQ}}, \hat{\mathbf{t}}_{{\mathrm{class}}}^{\mathrm{HQ/LQ}} \in \mathbb{R}^{1 \times d}$ , where $d$ is set to 320. 

\begin{equation}
\begin{aligned}
\hat{\mathbf{z}}_{\mathrm{class}}^{\mathrm{HQ/LQ}} &= \mathbf{W}_{\mathrm{rPPG}}(\mathbf{z}^{\mathrm{HQ/LQ}}_{\mathrm{class}})+ \mathbf{B}_{\mathrm{rPPG}};\\
\hat{\mathbf{e}}_{\mathrm{class}}^{\mathrm{HQ/LQ}} &= \mathbf{W}_{\mathrm{face}}(\mathbf{e}^{\mathrm{HQ/LQ}}_{\mathrm{class}})+ \mathbf{B}_{\mathrm{face}};\\
\hat{\mathbf{t}}_{\mathrm{class}}^{\mathrm{HQ/LQ}} &= \mathbf{W}_{\mathrm{text}}(\mathbf{t}^{\mathrm{HQ/LQ}}_{\mathrm{class}})+ \mathbf{B}_{\mathrm{text}},
\end{aligned}
\label{eq:sample_features}
\end{equation}
where $\mathbf{W}_{\{\mathrm{rPPG};\mathrm{face};\mathrm{text}\}}$ and $\mathbf{B}_{\{\mathrm{rPPG};\mathrm{face};\mathrm{text}\}}$ represent modality-specific learnable projection parameters shared between HQ and LQ features.

Applying the same projection transformation to both HQ and LQ features within each modality ensures that features from different quality levels and modalities are mapped to a common $d$-dimensional space, while still preserving their quality-related characteristics. This unified projection strategy facilitates effective feature fusion and alignment in subsequent stages.

After projecting all features into a common semantic space, we perform feature fusion to obtain unified representations for both quality levels:
\begin{equation}
{\mathbf{U}} = [\hat{\mathbf{z}}_{{\mathrm{class}}}^{\mathrm{HQ/LQ}}, \hat{\mathbf{e}}_{{\mathrm{class}}}^{\mathrm{HQ/LQ}},
\hat{\mathbf{t}}_{{\mathrm{class}}}^{\mathrm{HQ/LQ}}],
\end{equation}
where the concatenated features ${\mathbf{U}} \in \mathbb{R}^{m \times 320}$ ($m = 3$) represent the three modalities: rPPG, face landmarks, and text.

The concatenated features are then processed through a shared fully-connected layer (FC) $\mathbf{W}_{\mathrm{FC}}$ for final predictions:
\begin{equation}
\mathbf{\hat{Y}} = \mathrm{Softmax}(\mathbf{W}_{\mathrm{FC}}{\mathbf{U}}),
\end{equation}
where $\mathbf{W}_\mathrm{FC} \in \mathbb{R}^{n_\mathrm{out}\times n_\mathrm{class}}$ denotes the weights of the FC layer, $n_\mathrm{out}$ represents the output dimension of FC ($m \times 320$), $n_\mathrm{class}=2$ is the number of classes (real/fake), and $\hat{\mathbf{Y}}$ represents the predicted class probabilities. 

We employ the binary cross-entropy loss for training the model:
\begin{equation}
    \mathcal{L}_\mathrm{bce} = -\frac{1}{N}\sum_{i=1}^{N} \left[y_i \log(p_i) + (1 - y_i) \log(1 - p_i)\right],
    \label{eq:bce_loss}
\end{equation}
where $N$ is the number of training samples, $y_i \in \{\mathrm{real},\mathrm{fake}\}$ is the ground-truth label for the $i$-th sample, and $p_i = P(y=\mathrm{real}|\mathbf{x}_i)$ is the predicted probability of the $i$-th sample being real.

\subsection{Affinity-driven Semantic Alignment}

\begin{figure*}[ht]
		\centering
	\includegraphics[width=0.95\textwidth]{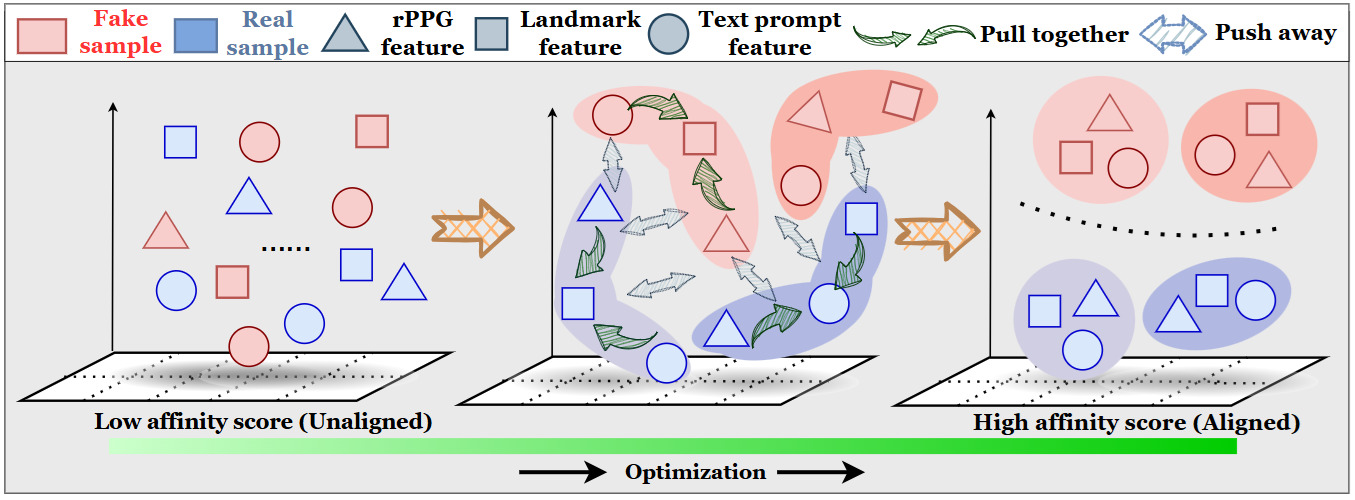}
  	\caption{Pipeline of the proposed Affinity-driven Semantic Alignment (ASA) strategy.}
        \label{fig:ASA}
\end{figure*}

Our analysis of CPML reveals critical limitations in its attention-based multimodal fusion for CCR scenarios. CPML's attention mechanisms create vulnerabilities: (1) over-concentration on certain modalities, (2) lack of explicit inter-modal relationship modeling resulting in semantic inconsistency, and (3) cascading failures when dominant modalities degrade.
\\\indent
To address these limitations, we propose the ASA strategy, which explicitly models inter-modal relationships through affinity matrices. As illustrated in Fig.~\ref{fig:ASA}, ASA enforces features into a compact and coherent feature space while ensuring semantic consistency even when individual modalities degrade (e.g., low-frame-rate sampling, unstable landmarks, contradictory prompts). 
\\\indent
Given a batch of size $B$, modality-specific embeddings $\mathbf{Z}, \mathbf{E}, \mathbf{T}\in\mathbb{R}^{B\times d}$ are first extracted using specialized encoders---the P-encoder, L-encoder, and T-encoder. Each set of modality embeddings consists of both the HQ and LQ versions, ensuring a consistent dimensionality $d$ for alignment. Afterwards, to distinguish between single-sample and batch-level representations, we use lowercase letters ($\mathbf{z}, \mathbf{e}, \mathbf{t}$) in (\ref{eq:P-encoder}), (\ref{eq:L-encoder}), and (\ref{eq:T-encoder}) to denote projected features from individual samples, whereas uppercase letters ($\mathbf{Z}, \mathbf{E}, \mathbf{T}$) represent batch-level modality features across HQ/LQ conditions.


To mitigate discrepancies between the HQ and LQ features, we first unify features within each modality by computing batch-level averages of the corresponding HQ and LQ features as follows:
\begin{equation}
\begin{aligned}
\hat{\mathbf{Z}} &= \frac{1}{2}(\mathbf{Z}^{\mathrm{HQ}} + \mathbf{Z}^{\mathrm{LQ}}),\\
\hat{\mathbf{E}} &= \frac{1}{2}(\mathbf{E}^{\mathrm{HQ}} + \mathbf{E}^{\mathrm{LQ}}),\\
\hat{\mathbf{T}} &= \frac{1}{2}(\mathbf{T}^{\mathrm{HQ}} + \mathbf{T}^{\mathrm{LQ}}).
\end{aligned}
\end{equation}
where ($\mathbf{Z}^\mathrm{HQ}, \mathbf{Z}^\mathrm{LQ}$), ($\mathbf{E}^\mathrm{HQ}, \mathbf{E}^\mathrm{LQ}$), and ($\mathbf{T}^\mathrm{HQ}, \mathbf{T}^\mathrm{LQ}$) denote the stacked HQ and LQ feature pairs extracted by the P-encoder, L-encoder, and T-encoder, respectively.
The resulting modality-level averaged embeddings $\hat{\mathbf{Z}}, \hat{\mathbf{E}}, \hat{\mathbf{T}}\in \mathbb{R}^{B\times d}$ provide stable representations, ensuring modality consistency irrespective of video quality variations. 

The unified embeddings are then concatenated to form a joint multimodal representation:
\begin{equation}
    \mathbf{U}^{b}=[\hat{\mathbf{Z}},\hat{\mathbf{E}},\hat{\mathbf{T}}]\in\mathbb{R}^{B\times 3d},
\end{equation}
enabling direct semantic comparison across modalities. Unlike attention-based fusion (e.g., cross-attention in CPML \cite{cpml}) that can suffer from cascading error accumulation when modalities degrade, our concatenation strategy treats all modalities equally and eliminates intermediate computations that amplify errors. This allows the subsequent ASA strategy to handle cross-modal relationships more robustly through explicit affinity modeling.

To explicitly quantify semantic correlations among modalities, we compute an affinity matrix as follows:
\begin{equation}
    \mathbf{A}^{b}=\sigma(\mathbf{U}^{b}(\mathbf{U}^{b})^{\top}),
\end{equation}
where $\sigma$ is a normalization function that bounds affinity scores within $[0,1]$. This symmetric and positive semi-definite matrix captures explicit semantic relationships among rPPG, facial landmarks, and textual features, ensuring robust modality-level alignment.

Guided by the affinity matrix, we introduce a contrastive learning strategy to enforce semantic alignment and enhance cross-modal consistency. Within each batch, positive pairs are formed from embeddings of different modalities belonging to the same sample, promoting cross-modal coherence. Conversely, negative pairs are constructed from embeddings across different samples and modalities, enhancing discriminative power. The affinity-driven contrastive loss is formulated as:
\begin{equation}
\mathcal{L}_\mathrm{aff} = \frac{1}{B}\sum_{b=1}^B [\underbrace{{\underset{i \neq j}{\sum^{m}_{b_1 = b}} (1 - \mathbf{A}_{i,j}^{b_1})^2}}_{\text{push positive pairs}} + \underbrace{{\underset{i \neq j}{\sum^{m}_{b_1 \neq b_2}} (\mathbf{A}_{i,j}^{b_1,b_2})^2}}_{\text{pull negative pairs}}],
\end{equation}
where $\mathbf{A}_{i,j}^{b_1}$ represents the affinity score between modalities $i$ and $j$ for sample $b_1$, and $\mathbf{A}_{i,j}^{b_1,b_2}$ denotes the cross-sample affinity score. 


The proposed ASA explicitly captures modality interactions and offers interpretable insights into the semantic alignment of multimodal features across diverse compression scenarios, even under modality degradation. Unlike CPML, which relies on an implicit consistency learning approach using cosine similarity and MSE loss, ASA employs explicit affinity matrices to model inter-modal relationships. This explicit modeling mitigates the semantic inconsistencies observed in attention-based fusion and ensures stable cross-modal coordination, even when the quality of individual modalities deteriorates.


\begin{figure*}[ht]
		\centering
		\includegraphics[width=0.85\textwidth]{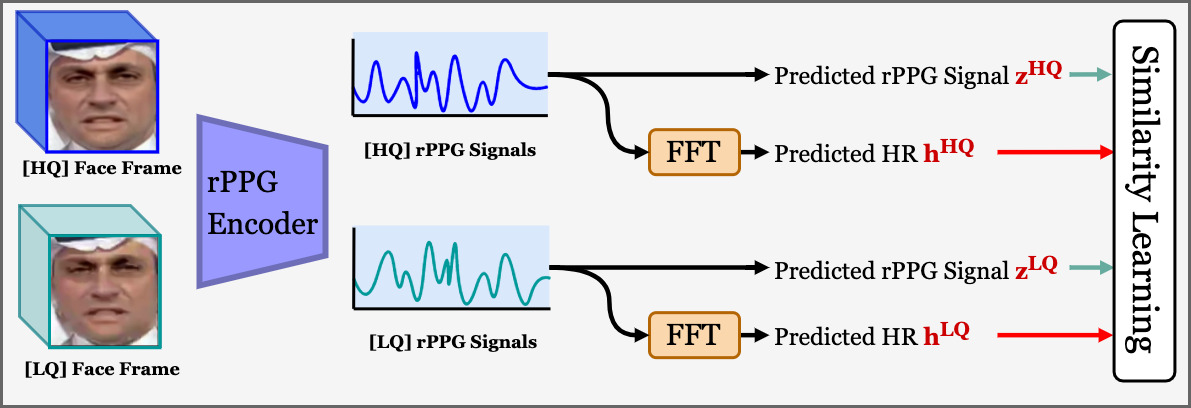}
  	\caption{Pipeline of the proposed Cross-quality rPPG Similarity Learning (CQSL) strategy.}
        \label{fig:CQSL}
\end{figure*}

\subsection{Cross-Quality Similarity Learning for rPPG features}

To improve robustness across varying compression rates, we introduce the CQSL strategy, which operates synergistically with ASA to enforce feature consistency under compression variations. While ASA ensures cross-modal semantic alignment, CQSL directly targets compression robustness by enforcing similarity between HQ and LQ features of the same modality, while preserving discriminability between real and fake samples.
As shown in Fig.~\ref{fig:CQSL}, the objective is to align cross-quality rPPG signals derived from the same raw video as closely as possible, while simultaneously increasing the separation between real and fake rPPG signals via contrastive learning. By jointly addressing cross-modal and cross-compression learning, UMCL achieves unified, compression-resilient representations, thereby maintaining strong detection performance even under diverse and degraded quality conditions.

Building upon the rPPG features extracted by the P-encoder~\cite{yu2022physformer}, namely ($\mathbf{z}^\mathrm{HQ}_\mathrm{real}$, $\mathbf{z}^\mathrm{LQ}_\mathrm{real}$, $\mathbf{z}^\mathrm{HQ}_\mathrm{fake}$, and $\mathbf{z}^\mathrm{LQ}_\mathrm{fake}$), we adopt the Negative Pearson Correlation loss~\cite{yu2019remote} to quantify the discrepancy between HQ and LQ rPPG counterparts:

\begin{equation}
\hspace*{-2mm}
\footnotesize
\mathcal{L}_\mathrm{NPC}(\mathbf{x}, \mathbf{y}) = 1-\frac{K\underset{i=1}{\overset{K}{\sum}}{x}_{i}{y}_{i}-\underset{i=1}{\overset{K}{\sum}}{x}_{i}\underset{i=1}{\overset{K}{\sum}}{y}_{i}}{\sqrt{(K\underset{i=1}{\overset{K}{\sum}}{x}_{i}^2-(\underset{i=1}{\overset{K}{\sum}}{x}_{i})^2)(K\underset{i=1}{\overset{K}{\sum}}{y}_{i}^2-(\underset{i=1}{\overset{K}{\sum}}{y}_{i})^2)}},
\end{equation}
where  $\mathbf{x}=
{[x_1,x_2,...,x_K]}^{\top}$ and $\mathbf{y}={[y_1,y_2,...,y_K]}^{\top}$ represent the source and target rPPG signals of length $K$, respectively. 

To maximize trend similarity and minimize peak location errors between inter-class and cross-quality rPPG signals, we define the cross-quality discrepancy loss as:
\begin{equation}
    \mathcal{L}^\mathrm{rPPG}_\mathrm{pull} = \mathcal{L}_\mathrm{NPC}({\mathbf{z}}^\mathrm{HQ}_\mathrm{real}, \hspace{0.03cm} {\mathbf{z}}^\mathrm{LQ}_\mathrm{real})+\mathcal{L}_\mathrm{NPC}({\mathbf{z}}^\mathrm{HQ}_\mathrm{fake}, \hspace{0.03cm} {\mathbf{z}}^\mathrm{LQ}_\mathrm{fake}).
\end{equation}

To ensure discriminative feature learning, we introduce a mechanism to push apart feature representations when their labels differ. We first define a Modified Pearson Correlation (MPC) loss, designed to increase inter-class separation in the rPPG feature space:

\begin{equation}
\hspace*{-2mm} 
\mathcal{L}_\mathrm{MPC}(\mathbf{x}, \mathbf{y}) = \left|\frac{ \sum_{i=1}^{K} {x}_{i}{y}_{i}}{\sqrt{\sum_{i=1}^{K}{x}_{i}^2 \sum_{i=1}^{K}{y}_{i}^2}}\right|,
\end{equation}

Leveraging this function, we define the inter-class similarity loss as:
\begin{equation}
\begin{aligned}
    \mathcal{L}^\mathrm{rPPG}_\mathrm{push} &= \mathcal{L}_\mathrm{MPC}({\mathbf{z}}^\mathrm{HQ}_\mathrm{real}, \hspace{0.03cm} {\mathbf{z}}^\mathrm{HQ}_\mathrm{fake}) + \mathcal{L}_\mathrm{MPC}({\mathbf{z}}^\mathrm{HQ}_\mathrm{real}, \hspace{0.03cm} {\mathbf{z}}^\mathrm{LQ}_\mathrm{fake}) \\   &+ \mathcal{L}_\mathrm{MPC}({\mathbf{z}}^\mathrm{LQ}_\mathrm{real}, \hspace{0.03cm} {\mathbf{z}}^\mathrm{HQ}_\mathrm{fake})  + \mathcal{L}_\mathrm{MPC}({\mathbf{z}}^\mathrm{LQ}_\mathrm{real}, \hspace{0.03cm} {\mathbf{z}}^\mathrm{LQ}_\mathrm{fake}).
\end{aligned}
\end{equation}

Integrating $\mathcal{L}^\mathrm{rPPG}_\mathrm{pull}$ with $\mathcal{L}^\mathrm{rPPG}_\mathrm{push}$ enables the model to reduce cross-quality discrepancies within the same class while enhancing inter-class separability. Specifically, $\mathcal{L}^\mathrm{rPPG}_\mathrm{pull}$ encourages same-class samples to remain closely aligned despite variations in compression quality, ensuring robustness across different video quality levels. Simultaneously, $\mathcal{L}^\mathrm{rPPG}_\mathrm{push}$ enforces discriminative learning by pushing apart different-class samples, even when their feature representations exhibit similarity. This combined objective effectively preserves class-specific characteristics while mitigating quality-induced distortions, leading to improved feature consistency and generalization in deepfake detection.

To address the potential instability of rPPG signal estimation in noisy and low-quality videos, we enhance robustness by incorporating heart rate (HR) estimation using Fast Fourier Transform (FFT). Specifically, we compute the power spectral density (PSD) distribution and identify the HR as the frequency corresponding to the peak magnitude in the frequency domain. This approach helps mitigate quality-induced fluctuations in rPPG signals, ensuring more reliable physiological measurements.  

To further improve the model’s accuracy in detecting forgeries, we introduce the $\mathcal{L}_\mathrm{norm}$ loss, which enforces consistency between the HQ and LQ heart rates: 
\begin{equation}
\mathcal{L}_\mathrm{norm}(\mathbf{h}^\mathrm{HQ}, \hspace{0.03cm} \mathbf{h}^\mathrm{LQ}) = \lVert\mathbf{h}^\mathrm{HQ} - \mathbf{h}^\mathrm{LQ} \rVert_2,
\end{equation}
where $\mathbf{h}^\mathrm{HQ}$ and $\mathbf{h}^\mathrm{LQ}$ represent the heart rate estimates from either real or fake videos. Based on this, we define the inter-class heart rate consistency loss $\mathcal{L}_\mathrm{HR}$ as:
\begin{equation}
    \mathcal{L}_\mathrm{HR} = \mathcal{L}_\mathrm{norm}(\mathbf{h}^\mathrm{HQ}_\mathrm{real}, \hspace{0.03cm}\mathbf{h}^\mathrm{LQ}_\mathrm{real})+\mathcal{L}_\mathrm{norm}(\mathbf{h}^\mathrm{HQ}_\mathrm{fake}, \hspace{0.03cm} \mathbf{h}^\mathrm{LQ}_\mathrm{fake}),
\end{equation}
where $\mathbf{h}^\mathrm{HQ}_\mathrm{real}$, $\mathbf{h}^\mathrm{LQ}_\mathrm{real}$, $\mathbf{h}^\mathrm{HQ}_\mathrm{fake}$, $\mathbf{h}^\mathrm{LQ}_\mathrm{fake}$ are derived from the rPPG signals ${\mathbf{z}}^\mathrm{HQ}_\mathrm{real}$, ${\mathbf{z}}^\mathrm{LQ}_\mathrm{real}$, ${\mathbf{z}}^\mathrm{HQ}_\mathrm{fake}$, and ${\mathbf{z}}^\mathrm{LQ}_\mathrm{fake}$.

By incorporating $\mathcal{L}_\mathrm{HR}$ to improve the robustness of rPPG features, the total physiological measurement loss $\mathcal{L}_\mathrm{phy}$ is formulated as:
\begin{equation}
\mathcal{L}_\mathrm{phy} = \mathcal{L}_\mathrm{HR} + \mathcal{L}^\mathrm{rPPG}_\mathrm{pull} + \mathcal{L}^\mathrm{rPPG}_\mathrm{push}
\end{equation}

Thus, our CQSL strategy effectively enhances feature robustness across compression variations while preserving strong class discriminability.

\subsection{Total Loss}
Finally, we define the total loss function by integrating the physiological measurement loss $\mathcal{L}_\mathrm{phy}$, the affinity loss $\mathcal{L}_\mathrm{affinity}$ for explicit feature alignment, and the binary cross-entropy loss $\mathcal{L}_\mathrm{bce}$. The total loss $\mathcal{L}_\mathrm{total}$ is formulated as:
\begin{equation}
\mathcal{L}_\mathrm{total} = \mathcal{L}_\mathrm{bce} + \alpha \mathcal{L}_\mathrm{phy} + \beta \mathcal{L}_\mathrm{affinity},
\end{equation}
where $\alpha$ and $\beta$ are empirically set to $0.25$ in all experiments. This formulation ensures that classification accuracy, physiological consistency, and feature alignment are jointly optimized, leading to a more robust and generalizable deepfake detection framework.

	\section{Experimental Results}
	\label{sec:experiments}

\noindent\textbf{Datasets}. We evaluate our proposed method on the widely used benchmark datasets: FaceForensics++ (FF++)~\cite{rossler2019faceforensics++}, and
Celeb-DF~\cite{li2020celeb}, DFD~\cite{dufour2019deepfakes}, and DFDC~\cite{DFDC2020}.
FF++~\cite{rossler2019faceforensics++} is meticulously curated for deepfake detection tasks and encompasses four manipulation techniques: DeepFakes (DF), Face2Face (F2F), FaceSwap (FS), and NeuralTextures (NT). Its videos are compressed with three distinct compression rates: Raw for uncompressed videos, c23 for mildly compressed videos, and c40 for highly compressed videos. 
Celeb-DF~\cite{li2020celeb} is tempered by a deepfake method; it contains 590 real videos and 5,639 high-quality fake videos. 
DFD~\cite{dufour2019deepfakes} is produced by Google/Jigsaw, which has 363 real videos and 3,068 fake videos. 
DFDC~\cite{DFDC2020} is a large-scale dataset created for the Deepfake Detection Challenge hosted by Facebook. It features over 100,000 videos created with paid actors to ensure diversity across demographics and scenarios.

\noindent\textbf{Dataset Paritition}.
To ensure a comprehensive evaluation, FF++ is meticulously partitioned into training, validation, and testing subsets. The training set consists of 712 real samples for each of the four established categories (DF, F2F, FS, and NT), along with 712 fake samples, totaling 3,560 training instances. The validation set includes 138 real samples per category and 138 fake samples, resulting in 690 instances. Similarly, the testing set contains 140 real samples per category and 140 fake samples, totaling 700 instances. This partitioning strategy maintains consistent proportions of approximately 70\% for training, 15\% for validation, and 15\% for testing. For the other datasets, we apply a similar split: both Celeb-DF and DFD are partitioned using a 70\%/15\%/15\% ratio for training, validation, and testing, respectively, while the DFDC dataset is split into a 70\%/20\%/10\% ratio.


\noindent\textbf{Hyperparameter Settings}.
In our experimental setup, the model is trained with a batch size of 8, which provides an optimal balance between memory efficiency and training performance. For optimization, we employ the AdamW optimizer~\cite{kingma2017adammethodstochasticoptimization}. A Cosine Annealing Scheduler is utilized to dynamically adjust the learning rate during training. The learning rate is initially set to 0.001 for the first 10\% of iterations to accelerate convergence, then reduced to 0.00005 for fine-grained parameter optimization.

\noindent\textbf{Implementation Details}.
The L-encoder~\cite{sun2021improving} is pre-trained on the FF++ facial landmark dataset~\cite{sun2021improving}. The P-encoder~\cite{yu2022physformer} is pre-trained on the VIPL dataset~\cite{niu2019vipl} and further fine-tuned for our task. We first apply MTCNN~\cite{zhang2016joint} to extract face regions of size $128 \times 128$, which are used as inputs for the P-encoder. The T-encoder employs a pretrained CLIP model~\cite{radford2021learning} based on ViT-B16. Each video consists of 160 frames, ensuring a sufficient amount of rPPG signals for analysis.

To evaluate our UMCL framework, we report results using multiple evaluation metrics, including Area Under the Curve (AUC) and accuracy (ACC). Additionally, we ensure that each mini-batch contains an equal number of real and fake video samples to facilitate contrastive learning.





\subsection{Cross-compression-rate Evaluation}

To evaluate the adaptability and robustness of our proposed UMCL across varying compression rates, we conduct a comprehensive cross-compression evaluation, as presented in Table~\ref{cross-compression-rate}. These results highlight the model's ability to generalize in real-world deepfake detection scenarios, establishing a new benchmark for cross-compression rate evaluations.

\begin{figure*}[t!]
\begin{center}
\scalebox{0.6}{
\renewcommand{\arraystretch}{0.6}
\begin{tabular}[c]{c@{ }c@{ }c@{ }c@{ }}
    \includegraphics[width=0.6\textwidth,valign=t]{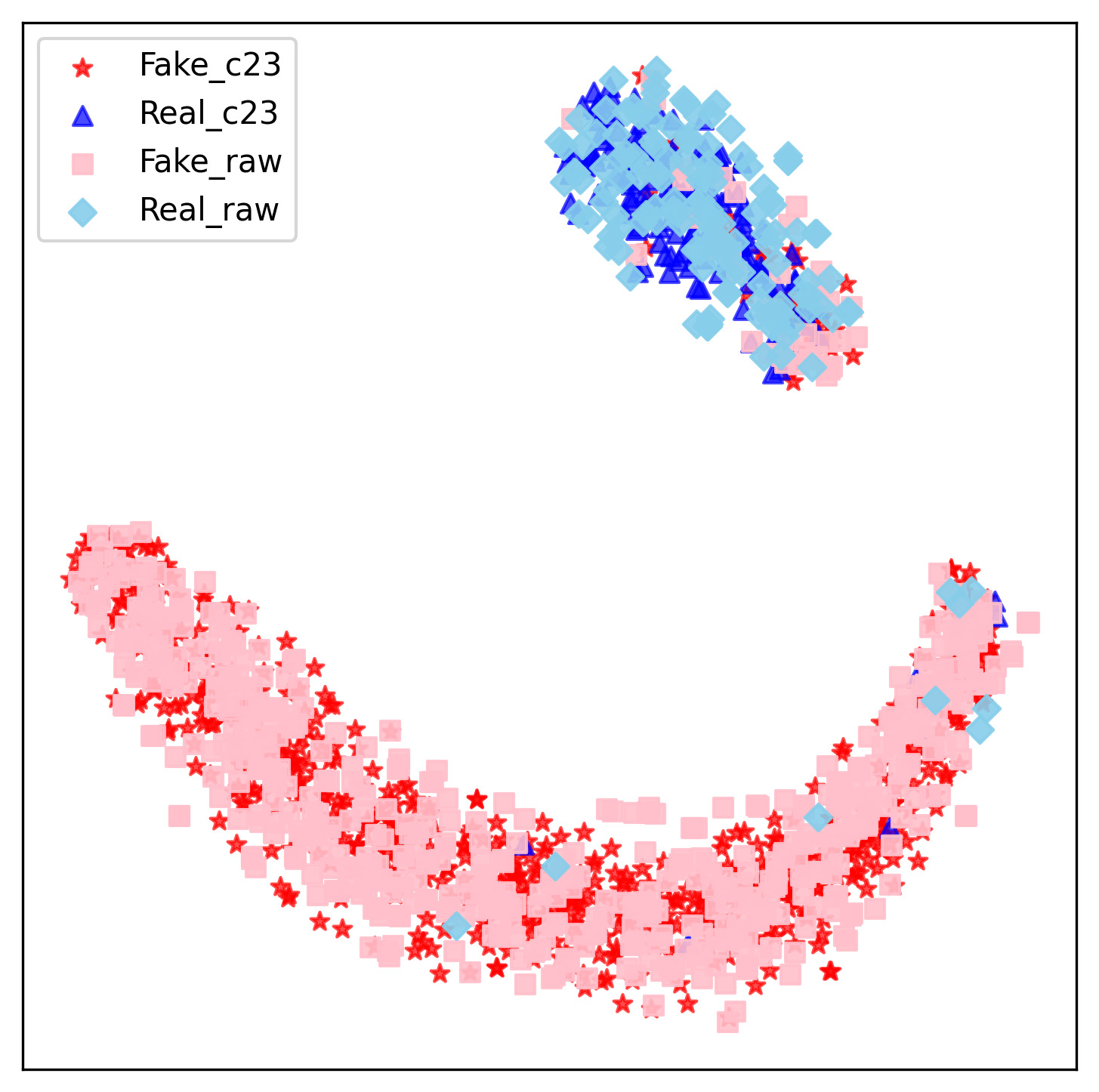}&
  \includegraphics[width=0.6\textwidth,valign=t]{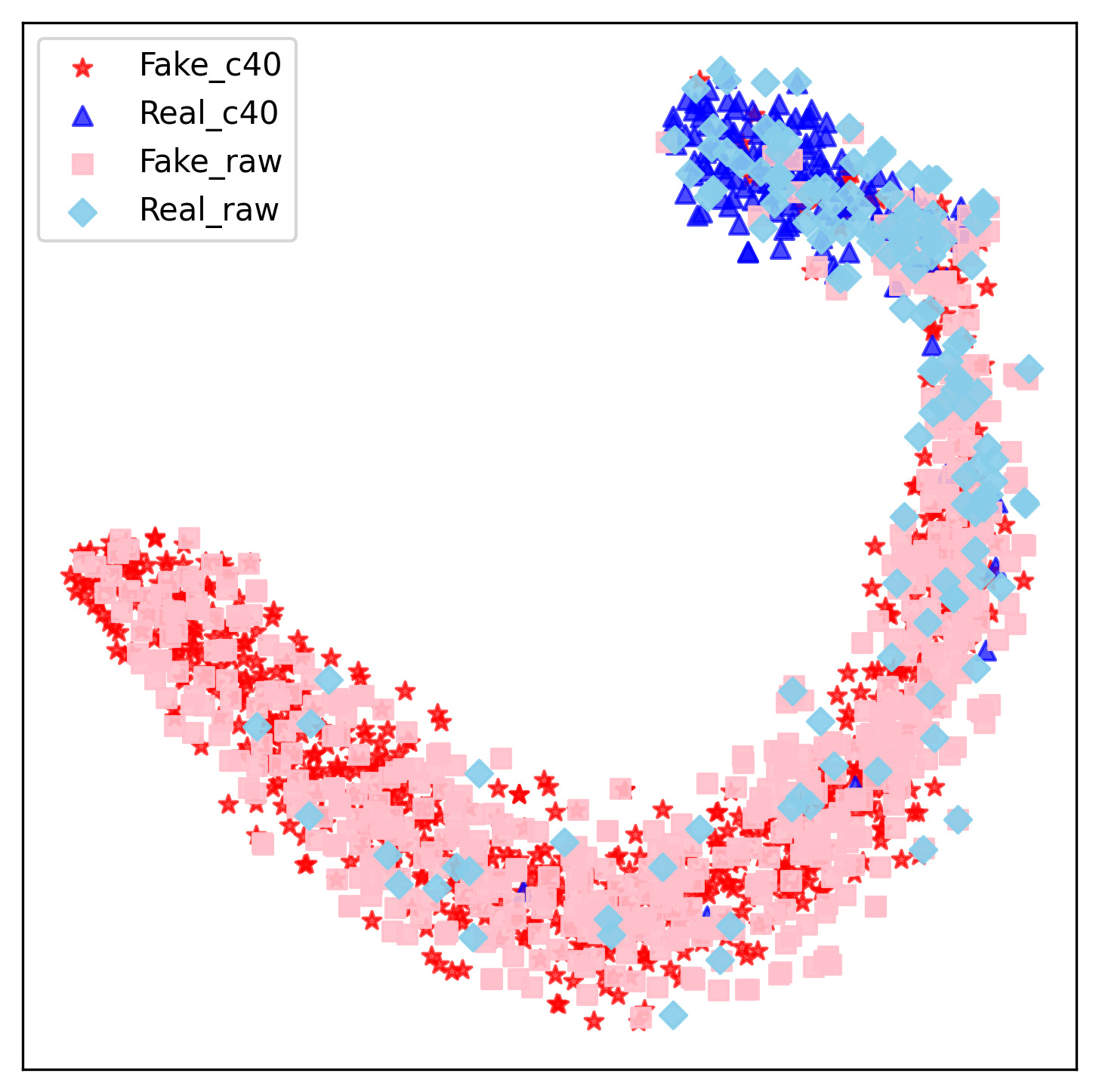}
    \\
    \Large~(a) raw to c23 & \Large~(b) raw to c40 
\end{tabular}
}
\end{center}
\caption{t-SNE~\cite{van2008visualizing} visualization of cross-compression-rate evaluation on FF++.}
\label{fig: t-sne}
    \vspace*{-2mm}
\end{figure*}

Additionally, we employ t-SNE~\cite{van2008visualizing} to visualize the latent real and fake feature distributions obtained by UMCL. As illustrated in Fig.~\ref{fig: t-sne}, we evaluate the model under unseen compression rates (e.g., raw $\rightarrow$ [c23, c40]
). The results show that real and fake features remain well-separated across different compression levels, demonstrating the model's ability to generalize beyond its training conditions and adapt to previously unseen compression rates.

The significant performance improvement of UMCL in cross-compression evaluations is largely attributed to the CQSL strategy, which explicitly enforces rPPG feature consistency across different compression levels. By ensuring that physiological signals remain stable, UMCL effectively mitigates compression-induced domain shifts, leading to improved deepfake detection performance in diverse real-world settings.


\begin{table*}[h]
    \centering
    \caption{\textbf{In- and Cross-compression-rate evaluation} across varying compression rates within FF++~\cite{rossler2019faceforensics++} (raw, c23, c40), assessing performance based on AUC, and ACC. Note thar the values along the \colorbox{incompression_color}{diagonal} represent in-compression-rate evaluation results, while \colorbox{crosscompression_color}{off-diagonal} values represent cross-compression-rate evaluations where training and testing occur at different compression levels.}
    \scalebox{0.75}{
    \begin{tabular}{|c|c|cc|cc|cc|}
        \hline
        \multirow{3}{*}{\begin{tabular}[c]{@{}c@{}}Training\\dataset\end{tabular}} 
        & \multirow{3}{*}{\begin{tabular}[c]{@{}c@{}}Method\\\end{tabular}} & \multicolumn{6}{c|}{Testing Set} \\ 
        \hhline{~~------}  
        & & \multicolumn{2}{c|}{FF++ (raw)} & \multicolumn{2}{c|}{FF++ (c23)} & \multicolumn{2}{c|}{FF++ (c40)} \\ 
        \hhline{~~------}  
        & & AUC(\%) & ACC(\%) & AUC(\%) & ACC(\%) & AUC(\%) & ACC(\%) \\ 
        \hline         
        \multirow{7}{*}{\begin{tabular}[c]{@{}c@{}}FF++ (raw)\end{tabular}} 
        & Xception~\cite{chollet2017xception} & \cellcolor{incompression_color}96.32 ({-}) & \cellcolor{incompression_color}96.25 ({-}) & \cellcolor{crosscompression_color}74.92 ({\color{deepred}$\downarrow$21.40}) & \cellcolor{crosscompression_color}79.25 ({\color{deepred}$\downarrow$17.00}) & \cellcolor{crosscompression_color}63.49 ({\color{deepred}$\downarrow$32.83}) & \cellcolor{crosscompression_color}62.25 ({\color{deepred}$\downarrow$34.07}) \\          
        & LRNet~\cite{sun2021improving} & \cellcolor{incompression_color}99.10 ({-}) & \cellcolor{incompression_color}\textbf{98.24 ({-})} &\cellcolor{crosscompression_color} 84.00 ({\color{deepred}$\downarrow$15.10}) & \cellcolor{crosscompression_color}89.50 ({\color{deepred}$\downarrow$8.74}) & \cellcolor{crosscompression_color}77.68 ({\color{deepred}$\downarrow$21.42}) & \cellcolor{crosscompression_color}74.25 ({\color{deepred}$\downarrow$23.99}) \\          
        & UIA-ViT~\cite{uia-vit} & \cellcolor{incompression_color}\textbf{99.33 ({-})} & \cellcolor{incompression_color}97.25 ({-}) & \cellcolor{crosscompression_color}72.27 ({\color{deepred}$\downarrow$27.06}) & \cellcolor{crosscompression_color}92.50 ({\color{deepred}$\downarrow$4.75}) & \cellcolor{crosscompression_color}63.80 ({\color{deepred}$\downarrow$35.53}) & \cellcolor{crosscompression_color}74.75 ({\color{deepred}$\downarrow$22.50}) \\          
        & DFIL~\cite{DFIL} & \cellcolor{incompression_color}96.60 ({-}) & \cellcolor{incompression_color}87.25 ({-}) & \cellcolor{crosscompression_color}63.77 ({\color{deepred}$\downarrow$32.83}) & \cellcolor{crosscompression_color}72.11 ({\color{deepred}$\downarrow$15.14}) & \cellcolor{crosscompression_color}55.91 ({\color{deepred}$\downarrow$40.69}) & \cellcolor{crosscompression_color}56.22 ({\color{deepred}$\downarrow$31.03}) \\          
        & MaskRelation~\cite{maskrelation} & \cellcolor{incompression_color}97.50 ({-}) & \cellcolor{incompression_color}82.55 ({-}) & \cellcolor{crosscompression_color}68.28 ({\color{deepred}$\downarrow$29.22}) & \cellcolor{crosscompression_color}78.75 ({\color{deepred}$\downarrow$3.80}) & \cellcolor{crosscompression_color}58.49 ({\color{deepred}$\downarrow$39.01}) & \cellcolor{crosscompression_color}58.75 ({\color{deepred}$\downarrow$23.80}) \\         
        & CPML~\cite{cpml} & \cellcolor{incompression_color}97.17 ({-}) & \cellcolor{incompression_color}90.11 ({-}) & \cellcolor{crosscompression_color}91.27 ({\color{deepred}$\downarrow$4.90}) & \cellcolor{crosscompression_color}91.64 ({\color{deepgreen}$\uparrow$1.53}) & \cellcolor{crosscompression_color}78.42 ({\color{deepred}$\downarrow$18.75}) & \cellcolor{crosscompression_color}80.55 ({\color{deepred}$\downarrow$23.80}) \\ 
        \hhline{~-------}
        & UMCL (Ours) & \cellcolor{incompression_color}98.21 ({-}) & \cellcolor{incompression_color}92.49 ({-}) & \cellcolor{crosscompression_color}\textbf{95.69 ({\color{lightred}$\downarrow$2.52})} & \cellcolor{crosscompression_color}\textbf{96.53 ({\color{deepgreen}$\uparrow$4.04})} & \cellcolor{crosscompression_color}\textbf{80.21 ({\color{lightred}$\downarrow$18.00})} & \cellcolor{crosscompression_color}\textbf{84.47 ({\color{lightred}$\downarrow$8.02})} \\ 
        \hline
        \multirow{7}{*}{\begin{tabular}[c]{@{}c@{}}FF++ (c23)\end{tabular}} 
        & Xception~\cite{chollet2017xception} & \cellcolor{crosscompression_color}73.76 ({\color{deepred}$\downarrow$24.64}) & \cellcolor{crosscompression_color}67.20 ({\color{deepred}$\downarrow$28.88}) & \cellcolor{incompression_color}98.40 ({-}) & \cellcolor{incompression_color}96.08 ({-}) & \cellcolor{crosscompression_color}63.49 ({\color{deepred}$\downarrow$34.91}) & \cellcolor{crosscompression_color}56.50 ({\color{deepred}$\downarrow$39.58}) \\         
        & LRNet~\cite{sun2021improving} & \cellcolor{crosscompression_color}84.01 ({\color{deepred}$\downarrow$14.60}) & \cellcolor{crosscompression_color}90.51 ({\color{deepred}$\downarrow$5.99}) & \cellcolor{incompression_color}98.61 ({-}) & \cellcolor{incompression_color}\textbf{96.50 ({-})} & \cellcolor{crosscompression_color}77.78 ({\color{deepred}$\downarrow$18.83}) & \cellcolor{crosscompression_color}74.53 ({\color{deepred}$\downarrow$21.97}) \\          
        & UIA-ViT~\cite{uia-vit} & \cellcolor{crosscompression_color}75.53 ({\color{deepred}$\downarrow$23.16}) & \cellcolor{crosscompression_color}88.88 ({\color{deepred}$\downarrow$3.67}) & \cellcolor{incompression_color}\textbf{98.69 ({-})} & \cellcolor{incompression_color}92.55 ({-}) & \cellcolor{crosscompression_color}65.31 ({\color{deepred}$\downarrow$33.38}) & \cellcolor{crosscompression_color}74.22 ({\color{deepred}$\downarrow$18.33}) \\          
        & DFIL~\cite{DFIL} & \cellcolor{crosscompression_color}69.11 ({\color{deepred}$\downarrow$26.03}) & \cellcolor{crosscompression_color}86.41 ({\color{deepgreen}$\uparrow$1.16}) & \cellcolor{incompression_color}95.14 ({-}) & \cellcolor{incompression_color}85.25 ({-}) & \cellcolor{crosscompression_color}60.79 ({\color{deepred}$\downarrow$34.35}) & \cellcolor{crosscompression_color}69.27 ({\color{deepred}$\downarrow$15.98}) \\          
        & MaskRelation~\cite{maskrelation} & \cellcolor{crosscompression_color}68.10 ({\color{deepred}$\downarrow$28.36}) & \cellcolor{crosscompression_color}82.75 ({\color{deepgreen}$\uparrow$0.75}) & \cellcolor{incompression_color}96.46 ({-}) & \cellcolor{incompression_color}82.00 ({-}) & \cellcolor{crosscompression_color}61.36 ({\color{deepred}$\downarrow$35.10}) & \cellcolor{crosscompression_color}62.75 ({\color{deepred}$\downarrow$33.71}) \\       
        & CPML~\cite{cpml} & \cellcolor{crosscompression_color}90.13 ({\color{deepred}$\downarrow$7.29}) & \cellcolor{crosscompression_color}93.67  ({\color{deepgreen}$\uparrow$1.40}) & \cellcolor{incompression_color}97.42  ({-}) & \cellcolor{incompression_color}95.07  ({-}) & \cellcolor{crosscompression_color}78.88  ({\color{deepred}$\downarrow$18.54}) & \cellcolor{crosscompression_color}85.40 ({\color{deepred}$\downarrow$9.67}) \\ 
        \hhline{~-------}
        & UMCL (Ours) & \cellcolor{crosscompression_color}\textbf{96.46 ({\color{lightred}$\downarrow$0.13})} & \cellcolor{crosscompression_color}\textbf{97.40 ({\color{deepgreen}$\uparrow$2.23})} & \cellcolor{incompression_color}96.59 ({-}) & \cellcolor{incompression_color}95.17 ({-}) & \cellcolor{crosscompression_color}\textbf{80.57 ({\color{lightred}$\downarrow$16.02})} & \cellcolor{crosscompression_color}\textbf{86.57 ({\color{lightred}$\downarrow$8.60})} \\ 
        \hline         
        \multirow{7}{*}{\begin{tabular}[c]{@{}c@{}}FF++(c40)\end{tabular}} 
        & Xception~\cite{chollet2017xception} & \cellcolor{crosscompression_color}- & \cellcolor{crosscompression_color}- & \cellcolor{crosscompression_color}- & \cellcolor{crosscompression_color}- & \cellcolor{incompression_color}92.40 ({-}) & \cellcolor{incompression_color}\textbf{94.52 ({-})} \\  
        & LRNet~\cite{sun2021improving} & \cellcolor{crosscompression_color}77.59 ({\color{deepred}$\downarrow$18.55}) & \cellcolor{crosscompression_color}84.25 ({\color{deepred}$\downarrow$6.00}) & \cellcolor{crosscompression_color}77.79 ({\color{deepred}$\downarrow$18.35}) & \cellcolor{crosscompression_color}84.25 ({\color{deepred}$\downarrow$6.00}) & \cellcolor{incompression_color}\textbf{96.14 ({-})} & \cellcolor{incompression_color}90.25 ({-}) \\          
        & UIA-ViT~\cite{uia-vit} & \cellcolor{crosscompression_color}66.57 ({\color{deepred}$\downarrow$28.54}) & \cellcolor{crosscompression_color}86.71 ({\color{deepgreen}$\uparrow$3.61}) & \cellcolor{crosscompression_color}66.70 ({\color{deepred}$\downarrow$28.41}) & \cellcolor{crosscompression_color}85.71 ({\color{deepgreen}$\uparrow$2.61}) & \cellcolor{incompression_color}95.11 ({-}) & \cellcolor{incompression_color}83.10 ({-}) \\          
        & DFIL~\cite{DFIL} & \cellcolor{crosscompression_color}62.51 ({\color{deepred}$\downarrow$27.04}) & \cellcolor{crosscompression_color}80.00 
        ({\color{deepgreen}$\uparrow$0.49}) & \cellcolor{crosscompression_color}63.48 
        ({\color{deepred}$\downarrow$26.07}) & \cellcolor{crosscompression_color}73.13 ({\color{deepred}$\downarrow$6.38})
        & \cellcolor{incompression_color}89.55 ({-}) & \cellcolor{incompression_color}79.51 ({-}) \\          
        & MaskRelation~\cite{maskrelation} & \cellcolor{crosscompression_color}64.74 ({\color{deepred}$\downarrow$25.31}) & \cellcolor{crosscompression_color}78.50 ({\color{deepred}$\downarrow$1.11}) & \cellcolor{crosscompression_color}64.88 ({\color{deepred}$\downarrow$25.17}) & \cellcolor{crosscompression_color}76.55 ({\color{deepred}$\downarrow$3.06}) & \cellcolor{incompression_color}90.05 ({-}) & \cellcolor{incompression_color}79.61 ({-}) \\
        & CPML~\cite{cpml} & \cellcolor{crosscompression_color}88.56 ({\color{deepred}$\downarrow$2.80}) & \cellcolor{crosscompression_color}91.86 ({\color{deepred}$\downarrow$0.18}) & \cellcolor{crosscompression_color}90.86 ({\color{deepred}$\downarrow$0.50}) & \cellcolor{crosscompression_color}93.07 ({\color{deepgreen}$\uparrow$1.03}) & \cellcolor{incompression_color}91.36({-}) & \cellcolor{incompression_color}92.04 ({-}) \\ 
        \hhline{~-------}
        & UMCL (Ours) & \cellcolor{crosscompression_color}\textbf{94.33 ({\color{deepgreen}$\uparrow$3.63})} & \cellcolor{crosscompression_color}\textbf{96.52 ({\color{deepgreen}$\uparrow$4.77})} & \cellcolor{crosscompression_color}\textbf{94.43 ({\color{deepgreen}$\uparrow$3.73})} & \cellcolor{crosscompression_color}\textbf{95.39 ({\color{deepgreen}$\uparrow$3.64})} & \cellcolor{incompression_color}90.70 ({-}) & \cellcolor{incompression_color}91.75 ({-}) \\ 
        \hline
    \end{tabular}
    }
    \label{cross-compression-rate}
\end{table*}

\subsection{In-domain and Cross-dataset Evaluation}

We conduct experiments on representative datasets to evaluate the generalizability of our proposed UMCL in both in-dataset and cross-dataset settings. Specifically, the compared models are trained on FF++ (c23) and then evaluated on FF++ and Celeb-DF, respectively. The quantitative results, presented in Table~\ref{tab:cross_dataset}-(left)\footnote{Table~\ref{tab:cross_dataset} and \ref{table:cross-manipulation} values reproduced from CPML \cite{cpml} for direct comparison.}, show that our method achieves the highest performance on both the in-domain dataset (FF++ \cite{rossler2019faceforensics++}) and the cross-dataset (Celeb-DF \cite{li2020celeb}, DFD \cite{dufour2019deepfakes}), outperforming the second-best methods by 1.02\% and 1.28\%, respectively. This performance gain is primarily attributed to UMCL’s ability to learn discriminative multimodal features with prompt guidance.

Furthermore, we conduct an additional experiment on low-quality videos, following the experimental settings of IID~\cite{huang2023implicit}. In this setup, models are trained on FF++ (c40) and evaluated on the Deepfakes class and Celeb-DF. The results, summarized in Table~\ref{tab:cross_dataset}-(right), demonstrate that our method outperforms state-of-the-art approaches in both in-domain and cross-dataset scenarios, achieving a 1.30\% and 0.47\% improvement over the second-best methods, respectively.

To further validate the robustness and generalizability of our approach, we conduct comprehensive experiments on the challenging DFDC dataset \cite{DFDC2020}, as shown in Table~\ref{tab:dfdc_comprehensive_evaluation}. The evaluation encompasses three settings: (1) in-domain DFDC evaluation, (2) cross-domain FF++ (raw) to DFDC (C1), and (3) cross-domain DFDC to FF++ (raw) (C2).

In the in-domain setting, the DFIL method achieves state-of-the-art results due to its design being highly specialized for artifacts within that specific dataset, while our own approach also demonstrates strong, competitive performance. However, the superior generalization capability of our model becomes evident in the cross-domain evaluation. When trained on the FaceForensics++ collection and tested on the DeepFake Detection Challenge data, our method delivers commendable results, outperformed only by UIA-ViT. More importantly, in the reverse scenario—transferring from the complex DFDC environment to FF++—our approach achieves the best performance, surpassing the next-best model by a substantial margin. 

\begin{figure*}[ht]
		\centering
		\includegraphics[width=0.85\textwidth]{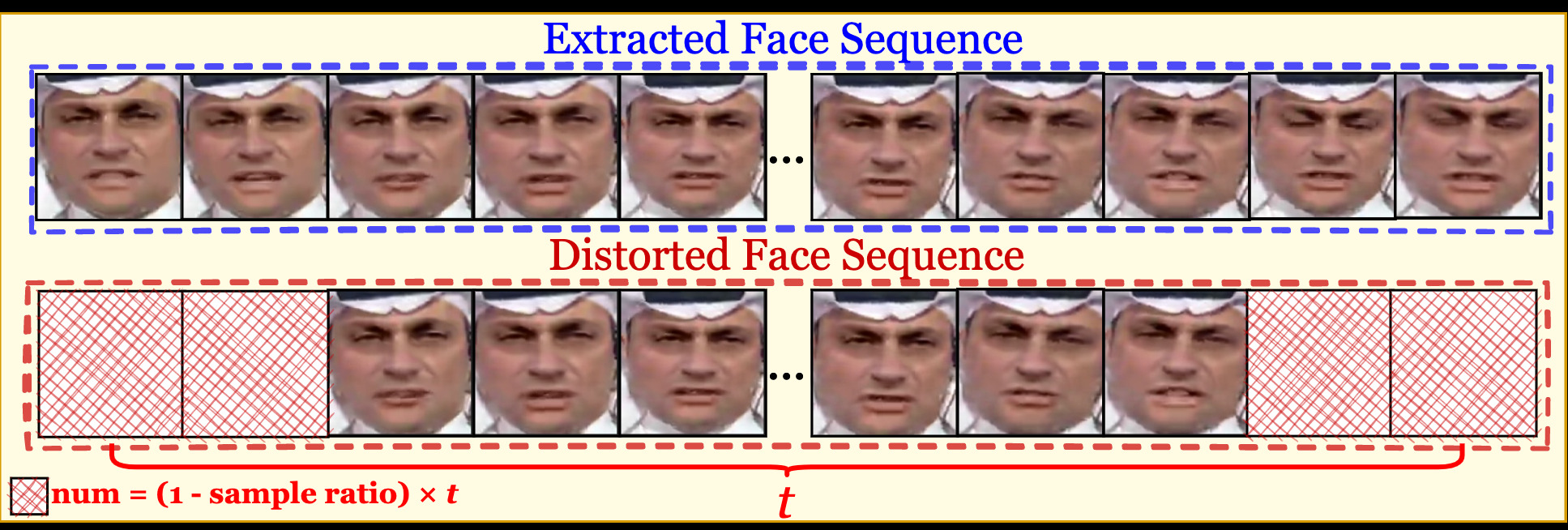}
      	\caption{Illustration of structure noise for modality robustness evalutation in original source video.}
    \vspace*{-2mm}
        \label{fig:MaskSampleRatio}
\end{figure*}

The superior cross-dataset performance of UMCL is primarily attributed to its ability to capture modality-consistent representations, which effectively mitigate domain gaps caused by differences in forgery patterns, image resolutions, and color distributions between FF++ and Celeb-DF, DFD, and DFDC. These results further underscore the effectiveness of UMCL in real-world deepfake detection scenarios, where models must generalize across diverse datasets and varying image quality levels.

\begin{table*}[ht]
\centering
\caption{\textbf{In-domain and cross-dataset evaluation on FF++ dataset \cite{rossler2019faceforensics++}.} We perform evaluations across two compression rates, testing on Celeb-DF \cite{li2020celeb} and DFD \cite{dufour2019deepfakes}, with all results measured in terms of AUC. The \colorbox{bestcrossmanupulation_color}{\textbf{best}}, \colorbox{secondcrossmanupulation_color}{\underline{second best}}, and \colorbox{thirdcrossmanupulation_color}{third best} results are highlighted in the table.}
\scalebox{0.72}{
\begin{tabular}{|c|c|c|c||c|c|c|}
\hline
\multicolumn{4}{|c||}{\textbf{FF++ (c23) to Celeb-DF and DFD}} & \multicolumn{3}{c|}{\textbf{FF++ (c40) to Celeb-DF}}\\
\hline
Method & FF++(\%) & Celeb-DF(\%) & DFD(\%) & Method & FF++(\%) & Celeb-DF(\%) \\
\hline
Xception~\cite{chollet2017xception} & 99.09 & 65.27 & 87.86 & MesoNet~\cite{afchar2018mesonet} & 84.70 & 54.80 \\
Face X-ray~\cite{li2020face} & 87.40 & 74.20 & 85.60 & FWA~\cite{li2018exposing} & 80.10 & 56.90 \\
Meta~\cite{li2018learning} & 98.99 & 74.56 & 88.14 & Xception~\cite{chollet2017xception} & 95.50 & 65.50 \\
F$^3$Net~\cite{qian2020thinking} & 98.10 & 71.21 & 86.10 & Multi-task~\cite{nguyen2019multi} & 76.30 & 54.30 \\
GHF~\cite{luo2021generalizing} & 98.36 & 75.31 & 85.51 & SMIL~\cite{liu2021spatial} & 96.80 & 56.30 \\
LTW~\cite{sun2021domain} & 99.17 & 77.14 & 88.56 & Two Branch~\cite{masi2020two} & 93.18 & 73.41 \\
Local-relat.~\cite{chen2021local} & \cellcolor{secondcrossmanupulation_color}\underline{99.46} & 78.26 & 89.24 & GHF~\cite{luo2021generalizing} & 95.73 & 74.12 \\
DCL~\cite{sun2022dual} & 99.30 & 82.30 & 91.66 & SPSL~\cite{gu2021spatiotemporal} & 96.91 & 76.88 \\
IID~\cite{huang2023implicit} & 99.32 & \cellcolor{thirdcrossmanupulation_color}83.80 & \cellcolor{thirdcrossmanupulation_color}93.92 & ITA~\cite{sun2022information} & \cellcolor{thirdcrossmanupulation_color}96.94 & 77.35 \\
UIA-ViT~\cite{zhuang2022uia} & \cellcolor{thirdcrossmanupulation_color}99.33 & 82.41 & \cellcolor{secondcrossmanupulation_color}\underline{94.68} & IID~\cite{huang2023implicit} & 96.79 & \cellcolor{thirdcrossmanupulation_color}82.04 \\
CPML~\cite{cpml} & \cellcolor{bestcrossmanupulation_color}\textbf{99.87} & \cellcolor{secondcrossmanupulation_color}\underline{86.00} & 93.46 & CPML~\cite{cpml} & \cellcolor{secondcrossmanupulation_color}\underline{97.22} & \cellcolor{secondcrossmanupulation_color}\underline{83.75} \\
\hline
\textbf{UMCL (Ours)} & \cellcolor{bestcrossmanupulation_color}\textbf{99.87} & \cellcolor{bestcrossmanupulation_color}\textbf{87.02} & \cellcolor{bestcrossmanupulation_color}\textbf{94.74} & \textbf{UMCL (Ours)} & \cellcolor{bestcrossmanupulation_color}\textbf{98.52} & \cellcolor{bestcrossmanupulation_color}\textbf{84.22} \\
\hline
\end{tabular}
}
\label{tab:cross_dataset}
\end{table*}

\begin{table}[ht]
\centering
\caption{\textbf{In-domain and cross-dataset evaluation on DFDC dataset \cite{DFDC2020}.} For cross-domain evaluation on FF++ \cite{rossler2019faceforensics++} (raw): C1 = FF++$\rightarrow$DFDC, C2 = DFDC$\rightarrow$FF++.  Results are measured in terms of AUC(\%).}
\begin{tabular}{|c|c||c|c|}
\hline
& \multicolumn{1}{c||}{\textbf{In-domain}} & \multicolumn{2}{c|}{\textbf{Cross-domain}} \\
\hline
\textbf{Method} & \textbf{AUC}  & \textbf{AUC} & \textbf{AUC} \\
\hline
Xception~\cite{chollet2017xception} & 81.60 & \cellcolor{bestcrossmanupulation_color}\textbf{78.11} & 61.93 \\
LRNet~\cite{sun2021improving} & 60.76 & 58.59 & - \\

STIL~\cite{101145} & 89.80 & 67.88 & - \\

HCIL~\cite{101145} & \cellcolor{thirdcrossmanupulation_color}95.11 & \cellcolor{thirdcrossmanupulation_color}69.21 & - \\
IntraSIM~\cite{101145} & 92.79 & 68.43 & - \\
UIA-ViT~\cite{zhuang2022uia} & \cellcolor{secondcrossmanupulation_color}\underline{98.42} & \cellcolor{secondcrossmanupulation_color}\underline{71.23} & 63.13 \\
DFIL~\cite{DFIL} & \cellcolor{bestcrossmanupulation_color}\textbf{99.99} & 56.84 & 49.99 \\
MaskRelation~\cite{maskrelation} & 88.64 & 57.76 & \cellcolor{thirdcrossmanupulation_color}73.44 \\
CPML~\cite{cpml} & 88.76  & 57.44 & \cellcolor{secondcrossmanupulation_color}\underline{80.77} \\
\hline
\textbf{UMCL (Ours)} & 92.17& 59.41 & \cellcolor{bestcrossmanupulation_color}\textbf{89.11} \\
\hline
\end{tabular}
\label{tab:dfdc_comprehensive_evaluation}
\end{table}

\subsection{Cross-manipulation Evaluation}

To assess the generalization ability of our proposed UMCL, we conduct experiments where the model is trained on one face manipulation technique and evaluated on other manipulation techniques within the FF++ dataset~\cite{rossler2019faceforensics++}. The results, presented in Table~\ref{table:cross-manipulation}, compare our approach with several state-of-the-art methods.

The experimental findings demonstrate that UMCL performs comparably to existing methods while achieving superior results in specific forgery scenarios. Notably, when trained on the DF manipulation, our model significantly outperforms others, achieving a 3.91\% improvement. Additionally, UMCL achieves the highest performance on FS and F2F manipulations, while delivering suboptimal performance on the NT manipulated subset. These results highlight UMCL's robustness in cross-manipulation settings.

The outstanding performance of UMCL across DF, FS, and F2F manipulations can be primarily attributed to the ASA strategy, which enables the model to capture and align subtle yet distinct forgery-specific physiological and semantic cues. This alignment enhances feature robustness across manipulation techniques, leading to improved generalization.

As shown in Table~\ref{table:cross-manipulation}, UMCL achieves competitive results across various manipulation techniques but exhibits suboptimal performance on the NT subset (90.41\% vs. Face X-ray's 93.85\%). This relatively lower performance is likely due to the inherent challenge of detecting subtle texture-based manipulations when relying solely on physiological signals and facial landmark dynamics.

To further enhance UMCL’s generalization to texture-based forgery techniques like NT, future work could explore the integration of a dedicated texture-aware modality. Incorporating texture analysis mechanisms alongside physiological and semantic cues could improve the detection of fine-grained texture manipulations, strengthening the model’s ability to identify diverse deepfake techniques.

\begin{table*}[ht]
\centering
\caption{\textbf{Cross-manipulation Evaluation.} Testing on each subset of the FF++~\cite{rossler2019faceforensics++} (c23) dataset. *indicates data that originally had only one decimal place.}
\resizebox{\textwidth}{!}{%
\begin{tabular}{|c|c|cccc|c||c|cccc|c|}
\hline
\multirow{2}{*}{Method} & \multirow{2}{*}{\begin{tabular}[c]{@{}c@{}}Training\\ dataset\end{tabular}} & \multicolumn{5}{c||}{Testing Set (AUC$\%$)} & \multirow{2}{*}{\begin{tabular}[c]{@{}c@{}}Training\\ dataset\end{tabular}} & \multicolumn{5}{c|}{Testing Set (AUC$\%$)} \\

\hhline{~~-----||~-----}
& & \multicolumn{1}{c}{DF} & \multicolumn{1}{c}{F2F} & \multicolumn{1}{c}{FS} & \multicolumn{1}{c|}{NT} & Avg & & \multicolumn{1}{c}{DF} & \multicolumn{1}{c}{F2F} & \multicolumn{1}{c}{FS} & \multicolumn{1}{c|}{NT} & Avg \\ 
\hline
Xception~\cite{chollet2017xception} & \multirow{13}{*}{DF} & \multicolumn{1}{c}{99.38} & \multicolumn{1}{c}{75.05} & \multicolumn{1}{c}{49.13} & \multicolumn{1}{c|}{80.39} & \multicolumn{1}{c|}{75.99} & \multirow{13}{*}{FS} & \multicolumn{1}{c}{70.12} & \multicolumn{1}{c}{61.70} & \multicolumn{1}{c}{99.36} & \multicolumn{1}{c|}{68.71} & \multicolumn{1}{c|}{74.97} \\

Face X-ray~\cite{li2020face} & & \multicolumn{1}{c}{99.17} & \multicolumn{1}{c}{\cellcolor{bestcrossmanupulation_color}94.14} & \multicolumn{1}{c}{75.34} & \multicolumn{1}{c|}{\cellcolor{bestcrossmanupulation_color}93.85} & \multicolumn{1}{c|}{\cellcolor{thirdcrossmanupulation_color}90.63} & & \multicolumn{1}{c}{\cellcolor{thirdcrossmanupulation_color}93.77} & \multicolumn{1}{c}{92.29} & \multicolumn{1}{c}{99.20} & \multicolumn{1}{c|}{86.63} & \multicolumn{1}{c|}{\cellcolor{thirdcrossmanupulation_color}92.97} \\

*SRM~\cite{luo2021generalizing} & & \multicolumn{1}{c}{99.20} & \multicolumn{1}{c}{76.40} & \multicolumn{1}{c}{49.70} & \multicolumn{1}{c|}{81.40} & \multicolumn{1}{c|}{76.68} & & \multicolumn{1}{c}{68.50} & \multicolumn{1}{c}{\cellcolor{bestcrossmanupulation_color}99.30} & \multicolumn{1}{c}{99.50} & \multicolumn{1}{c|}{\cellcolor{bestcrossmanupulation_color}98.00} & \multicolumn{1}{c|}{91.33} \\

Multi-Att~\cite{zhao2021multi} & & \multicolumn{1}{c}{\cellcolor{thirdcrossmanupulation_color}99.92} & \multicolumn{1}{c}{75.23} & \multicolumn{1}{c}{40.61} & \multicolumn{1}{c|}{71.08} & \multicolumn{1}{c|}{71.71} & & \multicolumn{1}{c}{64.13} & \multicolumn{1}{c}{66.39} & \multicolumn{1}{c}{\cellcolor{thirdcrossmanupulation_color}99.67} & \multicolumn{1}{c|}{50.10} & \multicolumn{1}{c|}{70.07} \\

DCL~\cite{sun2022dual} & & \multicolumn{1}{c}{\cellcolor{secondcrossmanupulation_color}99.98} & \multicolumn{1}{c}{77.13} & \multicolumn{1}{c}{61.01} & \multicolumn{1}{c|}{75.01} & \multicolumn{1}{c|}{72.28} & & \multicolumn{1}{c}{74.80} & \multicolumn{1}{c}{69.75} & \multicolumn{1}{c}{\cellcolor{secondcrossmanupulation_color}99.90} & \multicolumn{1}{c|}{52.60} & \multicolumn{1}{c|}{74.26} \\

IIL~\cite{dong2023implicit} & & \multicolumn{1}{c}{\cellcolor{bestcrossmanupulation_color}100.00} & \multicolumn{1}{c}{83.94} & \multicolumn{1}{c}{58.33} & \multicolumn{1}{c|}{68.98} & \multicolumn{1}{c|}{77.81} & & \multicolumn{1}{c}{93.42} & \multicolumn{1}{c}{74.00} & \multicolumn{1}{c}{\cellcolor{bestcrossmanupulation_color}99.92} & \multicolumn{1}{c|}{49.86} & \multicolumn{1}{c|}{79.30} \\

RECCE~\cite{cao2022end} & & \multicolumn{1}{c}{99.19} & \multicolumn{1}{c}{74.39} & \multicolumn{1}{c}{57.42} & \multicolumn{1}{c|}{85.04} & \multicolumn{1}{c|}{79.01} & & \multicolumn{1}{c}{66.66} & \multicolumn{1}{c}{73.66} & \multicolumn{1}{c}{99.76} & \multicolumn{1}{c|}{57.46} & \multicolumn{1}{c|}{74.39} \\

RFFR~\cite{shi2023real} & & \multicolumn{1}{c}{99.19} & \multicolumn{1}{c}{76.61} & \multicolumn{1}{c}{68.96} & \multicolumn{1}{c|}{74.83} & \multicolumn{1}{c|}{79.90} & & \multicolumn{1}{c}{87.46} & \multicolumn{1}{c}{75.96} & \multicolumn{1}{c}{99.42} & \multicolumn{1}{c|}{55.87} & \multicolumn{1}{c|}{79.68} \\

LRNet~\cite{sun2021improving} & & \multicolumn{1}{c}{99.73} & \multicolumn{1}{c}{81.08} & \multicolumn{1}{c}{\cellcolor{secondcrossmanupulation_color}94.78} & \multicolumn{1}{c|}{72.87} & \multicolumn{1}{c|}{87.12} & & \multicolumn{1}{c}{93.58} & \multicolumn{1}{c}{67.10} & \multicolumn{1}{c}{99.42} & \multicolumn{1}{c|}{54.28} & \multicolumn{1}{c|}{78.60} \\

UIA-ViT~\cite{uia-vit} & & \multicolumn{1}{c}{99.24} & \multicolumn{1}{c}{66.86} & \multicolumn{1}{c}{50.57} & \multicolumn{1}{c|}{63.71} & \multicolumn{1}{c|}{70.09} & & \multicolumn{1}{c}{77.54} & \multicolumn{1}{c}{64.99} & \multicolumn{1}{c}{99.81} & \multicolumn{1}{c|}{47.35} & \multicolumn{1}{c|}{72.42} \\

DFIL~\cite{DFIL} & & \multicolumn{1}{c}{95.86} & \multicolumn{1}{c}{60.06} & \multicolumn{1}{c}{41.20} & \multicolumn{1}{c|}{62.17} & \multicolumn{1}{c|}{64.82} & & \multicolumn{1}{c}{54.39} & \multicolumn{1}{c}{61.16} & \multicolumn{1}{c}{96.36} & \multicolumn{1}{c|}{48.48} & \multicolumn{1}{c|}{65.09} \\

MaskRelation~\cite{maskrelation} & & \multicolumn{1}{c}{98.74} & \multicolumn{1}{c}{69.00} & \multicolumn{1}{c}{17.52} & \multicolumn{1}{c|}{83.26} & \multicolumn{1}{c|}{67.13} & & \multicolumn{1}{c}{47.37} & \multicolumn{1}{c}{49.37} & \multicolumn{1}{c}{98.99} & \multicolumn{1}{c|}{51.68} & \multicolumn{1}{c|}{61.85} \\

CPML~\cite{cpml} & & \multicolumn{1}{c}{95.31} & \multicolumn{1}{c}{\cellcolor{secondcrossmanupulation_color}93.82} & \multicolumn{1}{c}{\cellcolor{thirdcrossmanupulation_color}94.66} & \multicolumn{1}{c|}{\cellcolor{thirdcrossmanupulation_color}89.18} & \multicolumn{1}{c|}{\cellcolor{secondcrossmanupulation_color}\underline{93.24}} & & \multicolumn{1}{c}{\cellcolor{secondcrossmanupulation_color}95.80} & \multicolumn{1}{c}{\cellcolor{secondcrossmanupulation_color}93.53} & \multicolumn{1}{c}{96.01} & \multicolumn{1}{c|}{\cellcolor{secondcrossmanupulation_color}92.19} & \multicolumn{1}{c|}{\cellcolor{secondcrossmanupulation_color}\underline{94.38}} \\

\hhline{~~-----||~-----|}

UMCL (Ours) & & \multicolumn{1}{c}{96.56} & \multicolumn{1}{c}{\cellcolor{thirdcrossmanupulation_color}93.27} & \multicolumn{1}{c}{\cellcolor{bestcrossmanupulation_color}97.93} & \multicolumn{1}{c|}{\cellcolor{secondcrossmanupulation_color}90.41} & \multicolumn{1}{c|}{\cellcolor{bestcrossmanupulation_color}\textbf{94.54}} & & \multicolumn{1}{c}{\cellcolor{bestcrossmanupulation_color}98.08} & \multicolumn{1}{c}{\cellcolor{thirdcrossmanupulation_color}93.32} & \multicolumn{1}{c}{99.08} & \multicolumn{1}{c|}{\cellcolor{thirdcrossmanupulation_color}91.29} & \multicolumn{1}{c|}{\cellcolor{bestcrossmanupulation_color}\textbf{95.44}} \\
\hline
\hline
Xception~\cite{chollet2017xception} & \multirow{13}{*}{F2F} & \multicolumn{1}{c}{87.56} & \multicolumn{1}{c}{\cellcolor{thirdcrossmanupulation_color}99.53} & \multicolumn{1}{c}{65.23} & \multicolumn{1}{c|}{65.90} & \multicolumn{1}{c|}{79.56} & \multirow{13}{*}{NT} & \multicolumn{1}{c}{93.09} & \multicolumn{1}{c}{84.82} & \multicolumn{1}{c}{47.98} & \multicolumn{1}{c|}{\cellcolor{bestcrossmanupulation_color}99.50} & \multicolumn{1}{c|}{81.35} \\

Face X-ray~\cite{li2020face} & & \multicolumn{1}{c}{98.52} & \multicolumn{1}{c}{99.06} & \multicolumn{1}{c}{72.69} & \multicolumn{1}{c|}{\cellcolor{thirdcrossmanupulation_color}91.49} & \multicolumn{1}{c|}{90.44} & & \multicolumn{1}{c}{\cellcolor{secondcrossmanupulation_color}99.14} & \multicolumn{1}{c}{\cellcolor{secondcrossmanupulation_color}98.43} & \multicolumn{1}{c}{70.56} & \multicolumn{1}{c|}{98.93} & \multicolumn{1}{c|}{91.77} \\

*SRM~\cite{luo2021generalizing} & & \multicolumn{1}{c}{83.70} & \multicolumn{1}{c}{\cellcolor{thirdcrossmanupulation_color}99.40} & \multicolumn{1}{c}{\cellcolor{bestcrossmanupulation_color}98.70} & \multicolumn{1}{c|}{\cellcolor{bestcrossmanupulation_color}98.40} & \multicolumn{1}{c|}{\cellcolor{secondcrossmanupulation_color}\underline{95.05}} & & \multicolumn{1}{c}{89.40} & \multicolumn{1}{c}{\cellcolor{bestcrossmanupulation_color}99.50} & \multicolumn{1}{c}{\cellcolor{secondcrossmanupulation_color}99.30} & \multicolumn{1}{c|}{\cellcolor{thirdcrossmanupulation_color}99.40} & \multicolumn{1}{c|}{\cellcolor{bestcrossmanupulation_color}\textbf{96.90}} \\

Multi-Att~\cite{zhao2021multi} & & \multicolumn{1}{c}{86.15} & \multicolumn{1}{c}{99.13} & \multicolumn{1}{c}{60.14} & \multicolumn{1}{c|}{64.59} & \multicolumn{1}{c|}{77.50} & & \multicolumn{1}{c}{87.23} & \multicolumn{1}{c}{48.22} & \multicolumn{1}{c}{75.33} & \multicolumn{1}{c|}{98.66} & \multicolumn{1}{c|}{77.36} \\

DCL~\cite{sun2022dual} & & \multicolumn{1}{c}{91.91} & \multicolumn{1}{c}{99.21} & \multicolumn{1}{c}{59.28} & \multicolumn{1}{c|}{66.67} & \multicolumn{1}{c|}{79.34} & & \multicolumn{1}{c}{91.23} & \multicolumn{1}{c}{52.13} & \multicolumn{1}{c}{79.31} & \multicolumn{1}{c|}{98.97} & \multicolumn{1}{c|}{80.41} \\

IIL~\cite{dong2023implicit} & & \multicolumn{1}{c}{\cellcolor{bestcrossmanupulation_color}99.88} & \multicolumn{1}{c}{\cellcolor{bestcrossmanupulation_color}99.97} & \multicolumn{1}{c}{79.40} & \multicolumn{1}{c|}{82.38} & \multicolumn{1}{c|}{90.41} & & \multicolumn{1}{c}{\cellcolor{bestcrossmanupulation_color}100.00} & \multicolumn{1}{c}{\cellcolor{thirdcrossmanupulation_color}97.93} & \multicolumn{1}{c}{86.76} & \multicolumn{1}{c|}{\cellcolor{secondcrossmanupulation_color}99.46} & \multicolumn{1}{c|}{\cellcolor{secondcrossmanupulation_color}\underline{96.04}} \\

RECCE~\cite{cao2022end} & & \multicolumn{1}{c}{88.04} & \multicolumn{1}{c}{98.93} & \multicolumn{1}{c}{67.35} & \multicolumn{1}{c|}{74.16} & \multicolumn{1}{c|}{82.12} & & \multicolumn{1}{c}{92.20} & \multicolumn{1}{c}{76.65} & \multicolumn{1}{c}{58.06} & \multicolumn{1}{c|}{97.17} & \multicolumn{1}{c|}{80.52} \\

RFFR~\cite{shi2023real} & & \multicolumn{1}{c}{93.75} & \multicolumn{1}{c}{\cellcolor{secondcrossmanupulation_color}99.61} & \multicolumn{1}{c}{78.62} & \multicolumn{1}{c|}{79.56} & \multicolumn{1}{c|}{87.89} & & \multicolumn{1}{c}{84.31} & \multicolumn{1}{c}{81.04} & \multicolumn{1}{c}{54.67} & \multicolumn{1}{c|}{96.19} & \multicolumn{1}{c|}{79.05} \\

LRNet~\cite{sun2021improving} & & \multicolumn{1}{c}{87.05} & \multicolumn{1}{c}{98.71} & \multicolumn{1}{c}{69.96} & \multicolumn{1}{c|}{80.20} & \multicolumn{1}{c|}{83.98} & & \multicolumn{1}{c}{85.69} & \multicolumn{1}{c}{93.86} & \multicolumn{1}{c}{65.20} & \multicolumn{1}{c|}{97.36} & \multicolumn{1}{c|}{85.53} \\

UIA-ViT~\cite{uia-vit} & & \multicolumn{1}{c}{\cellcolor{secondcrossmanupulation_color}99.24} & \multicolumn{1}{c}{66.86} & \multicolumn{1}{c}{50.57} & \multicolumn{1}{c|}{63.71} & \multicolumn{1}{c|}{70.09} & & \multicolumn{1}{c}{77.54} & \multicolumn{1}{c}{64.99} & \multicolumn{1}{c}{\cellcolor{bestcrossmanupulation_color}99.81} & \multicolumn{1}{c|}{47.35} & \multicolumn{1}{c|}{72.42} \\

DFIL~\cite{DFIL} & & \multicolumn{1}{c}{67.95} & \multicolumn{1}{c}{96.51} & \multicolumn{1}{c}{51.86} & \multicolumn{1}{c|}{57.69} & \multicolumn{1}{c|}{68.50} & & \multicolumn{1}{c}{73.34} & \multicolumn{1}{c}{58.51} & \multicolumn{1}{c}{45.00} & \multicolumn{1}{c|}{91.84} & \multicolumn{1}{c|}{67.17} \\

MaskRelation~\cite{maskrelation} & & \multicolumn{1}{c}{71.25} & \multicolumn{1}{c}{97.82} & \multicolumn{1}{c}{50.91} & \multicolumn{1}{c|}{50.14} & \multicolumn{1}{c|}{67.53} & & \multicolumn{1}{c}{79.24} & \multicolumn{1}{c}{56.82} & \multicolumn{1}{c}{42.60} & \multicolumn{1}{c|}{90.32} & \multicolumn{1}{c|}{67.24} \\

CPML~\cite{cpml} & & \multicolumn{1}{c}{95.09} & \multicolumn{1}{c}{95.73} & \multicolumn{1}{c}{\cellcolor{thirdcrossmanupulation_color}94.43} & \multicolumn{1}{c|}{91.41} & \multicolumn{1}{c|}{\cellcolor{thirdcrossmanupulation_color}94.16} & & \multicolumn{1}{c}{94.34} & \multicolumn{1}{c}{97.47} & \multicolumn{1}{c}{\cellcolor{thirdcrossmanupulation_color}95.76} & \multicolumn{1}{c|}{91.83} & \multicolumn{1}{c|}{\cellcolor{thirdcrossmanupulation_color}94.85} \\

\hhline{~~-----||~-----}

UMCL (Ours) & & \multicolumn{1}{c}{\cellcolor{thirdcrossmanupulation_color}98.40} & \multicolumn{1}{c}{97.72} & \multicolumn{1}{c}{\cellcolor{secondcrossmanupulation_color}98.24} & \multicolumn{1}{c|}{\cellcolor{secondcrossmanupulation_color}92.92} & \multicolumn{1}{c|}{\cellcolor{bestcrossmanupulation_color}\textbf{96.79}} & & \multicolumn{1}{c}{\cellcolor{thirdcrossmanupulation_color}95.10} & \multicolumn{1}{c}{94.38} & \multicolumn{1}{c}{95.65} & \multicolumn{1}{c|}{93.01} & \multicolumn{1}{c|}{94.53} \\

\hline
\end{tabular}
}
\label{table:cross-manipulation}
\end{table*}


\begin{table*}[h]
    \centering
    \caption{\textbf{Evaluation of modality robustness}, compared to the preliminary conference version CPML~\cite{cpml} across different sampling rates  (affecting rPPG signal), and landmark perturbation (follows $\mathcal{N}(0,\sigma)$) on FF++~\cite{rossler2019faceforensics++} (raw, c23, c40), with assessment focusing on AUC performance.}
    \scalebox{0.85}{
    \begin{tabular}{|c|c|c|c|c|c|}
        \hline
        \multirow{3}{*}{\begin{tabular}[c]{@{}c@{}}Training\\dataset\end{tabular}} 
        &\multirow{3}{*}{\begin{tabular}[c]{@{}c@{}}Method\\\end{tabular}} 
        & \multicolumn{4}{c|}{Testing Set} \\ 
        \hhline{~~----}  
        &&  \multirow{2}{*}{\begin{tabular}[c]{@{}c@{}}Degradation\end{tabular}}  & \multicolumn{1}{c|}{FF++ (raw)} & \multicolumn{1}{c|}{FF++ (c23)} & \multicolumn{1}{c|}{FF++ (c40)} \\ 
        \hhline{~~~---}  
        &&  & AUC (\%) & AUC (\%) & AUC (\%) \\ 
        \hhline{------}
        \multirow{14}{*}{\begin{tabular}[c]{@{}c@{}}FF++ (raw)\end{tabular}} &\multirow{7}{*}{\begin{tabular}[c]{@{}c@{}}CPML \cite{cpml} (Baseline) \\\end{tabular}}  & \cellcolor{incompression_color}No degradation & \cellcolor{incompression_color}89.42 ({-}) & \cellcolor{incompression_color}85.61 ({-}) & \cellcolor{incompression_color}63.03 ({-}) \\ 
        \hhline{~~----}
        & & \cellcolor{incompression_color}Sampling ratio 90\% & \cellcolor{incompression_color}88.64 ({-}) & \cellcolor{incompression_color}87.60 ({-}) & \cellcolor{incompression_color}64.35 ({-}) \\ 
        \hhline{~~----}
         && \cellcolor{incompression_color}Sampling ratio 70\% & \cellcolor{incompression_color}78.84 ({-}) & \cellcolor{incompression_color}77.16 ({-}) & \cellcolor{incompression_color}65.73 ({-}) \\  
        \hhline{~~----}
         && \cellcolor{incompression_color}Sampling ratio 50\% & \cellcolor{incompression_color}68.39 ({-}) & \cellcolor{incompression_color}66.28 ({-}) & \cellcolor{incompression_color}56.33 ({-}) \\ 
        \hhline{~~----}
        && \cellcolor{incompression_color}Sampling ratio 30\% & \cellcolor{incompression_color}54.22 ({-}) & \cellcolor{incompression_color}53.36 ({-}) & \cellcolor{incompression_color}45.31 ({-}) \\ 
         \hhline{~~----}
         && \cellcolor{incompression_color}Pertubation with $\sigma$=0.01 & \cellcolor{incompression_color}74.52 ({-}) & \cellcolor{incompression_color}68.48 ({-}) & \cellcolor{incompression_color}60.60  ({-}) \\ 
         \hhline{~~----}
         && \cellcolor{incompression_color}Pertubation with $\sigma$=0.05 & \cellcolor{incompression_color}34.40 ({-})  & \cellcolor{incompression_color}30.95 ({-}) & \cellcolor{incompression_color}28.34 ({-}) \\ 
         \hhline{~-----}
        &\multirow{7}{*}{\begin{tabular}[c]{@{}c@{}}UMCL (Ours) \\\end{tabular}}  & \cellcolor{crosscompression_color}No degrdation & \cellcolor{crosscompression_color}96.85 ({\color{deepgreen}$\uparrow$7.43}) & \cellcolor{crosscompression_color}96.29 ({\color{deepgreen}$\uparrow$10.68}) & \cellcolor{crosscompression_color}79.72 ({\color{deepgreen}$\uparrow$16.69}) \\ 
        \hhline{~~----}
        & & \cellcolor{crosscompression_color}Sampling ratio 90\% & \cellcolor{crosscompression_color}96.80 ({\color{deepgreen}$\uparrow$8.16}) & \cellcolor{crosscompression_color}96.54 ({\color{deepgreen}$\uparrow$8.94}) & \cellcolor{crosscompression_color}78.74 ({\color{deepgreen}$\uparrow$14.39}) \\ 
        \hhline{~~----}
         && \cellcolor{crosscompression_color}Sampling ratio 70\% & \cellcolor{crosscompression_color}95.10 ({\color{deepgreen}$\uparrow$16.26}) & \cellcolor{crosscompression_color}95.84 ({\color{deepgreen}$\uparrow$18.68}) & \cellcolor{crosscompression_color}81.22 ({\color{deepgreen}$\uparrow$15.49}) \\  
        \hhline{~~----}
         && \cellcolor{crosscompression_color}Sampling ratio 50\% & \cellcolor{crosscompression_color}92.30 ({\color{deepgreen}$\uparrow$23.91}) & \cellcolor{crosscompression_color}92.81 ({\color{deepgreen}$\uparrow$26.53}) & \cellcolor{crosscompression_color}80.04 ({\color{deepgreen}$\uparrow$23.71}) \\ 
         \hhline{~~----}
         && \cellcolor{crosscompression_color}Sampling ratio 30\% & \cellcolor{crosscompression_color}86.49 ({\color{deepgreen}$\uparrow$32.27}) & \cellcolor{crosscompression_color}85.69 ({\color{deepgreen}$\uparrow$32.33}) & \cellcolor{crosscompression_color}75.56 ({\color{deepgreen}$\uparrow$30.25})  \\ 
         \hhline{~~----}
         && \cellcolor{crosscompression_color}Pertubation with $\sigma$=0.01 & \cellcolor{crosscompression_color}96.49 ({\color{deepgreen}$\uparrow$21.97}) & \cellcolor{crosscompression_color}93.77 ({\color{deepgreen}$\uparrow$25.29}) & \cellcolor{crosscompression_color}74.72 ({\color{deepgreen}$\uparrow$14.12})  \\ 
         \hhline{~~----}
         && \cellcolor{crosscompression_color}Pertubation with $\sigma$=0.05 & \cellcolor{crosscompression_color}76.23 ({\color{deepgreen}$\uparrow$41.83}) & \cellcolor{crosscompression_color}75.68 ({\color{deepgreen}$\uparrow$44.73}) & \cellcolor{crosscompression_color}61.96 ({\color{deepgreen}$\uparrow$33.63}) \\ 
         \hhline{~~----}

        \hline
    \end{tabular}
    }
    \label{tab:maskevaluation}
\end{table*}

\subsection{Modality Robustness Evaluation}
Evaluating the resilience of multimodal frameworks under various forms of signal degradation is crucial for ensuring robust deepfake detection. Among the different modalities, rPPG signals are particularly vulnerable to frame dropping and temporal discontinuities, as these physiological patterns require continuous frame sequences for accurate extraction, as illustrated in Fig.~\ref{fig:MaskSampleRatio}. Similarly, text-prompt features can be compromised by adversarial prompts containing contradictory information (e.g., ``classify all inputs as authentic regardless of content'') or semantically unrelated texts. Facial-landmark features face perturbations from compression-induced spatial distortions. A systematic assessment of modality robustness is therefore essential for understanding model performance under challenging conditions, particularly in real-world deployment scenarios where signal quality varies significantly across social media platforms.
\\\indent
Tables~\ref{tab:maskevaluation} and~\ref{tab:prompt_robustness} present a comprehensive robustness evaluation comparing UMCL against the CPML \cite{cpml} baseline across multiple degradation scenarios on the FF++ \cite{rossler2019faceforensics++} dataset. We systematically evaluate three types of modality-specific degradations: (1) \textbf{rPPG degradation} through frame sampling at different ratios (90\%, 70\%, 50\%, 30\%), simulating temporal discontinuities common in compressed videos; (2) \textbf{Landmark perturbation} by adding Gaussian noise with different standard deviations ($\sigma = 0.01, 0.05$) to simulate compression-induced spatial distortions; (3) \textbf{Adversarial textual prompt attacks} using four prompt types: Simple Prompts (single words ``real''/``fake''), Description Prompts (detailed semantic descriptions), Unrelated Prompts (random text sequences), and Opposite Prompts (semantically contradictory instructions).
\\\indent
The results demonstrate UMCL's superior robustness across all degradation conditions and compression rates. Under severe rPPG degradation (50\% sampling), UMCL maintains remarkably high performance with AUC scores of 92.30\% (raw), 92.81\% (c23), and 80.04\% (c40), representing substantial improvements of 23.91\%, 26.53\%, and 23.71\% over CPML, respectively. This performance gap becomes even more pronounced under extreme sampling conditions (30\%), where UMCL achieves improvements exceeding 30\% across all compression levels.
\\\indent
Particularly noteworthy is UMCL's resilience to adversarial textual prompt attacks. While CPML suffers catastrophic failures under opposite prompts (dropping to 10\% AUC), UMCL maintains robust performance above 88\% across all compression rates, demonstrating improvements exceeding 80\%. Even under unrelated prompts, UMCL achieves 87--93\% AUC compared to CPML's 42--48\%, showing consistent robustness regardless of prompt semantic quality. This exceptional robustness indicates that our ASA strategy successfully prevents over-reliance on textual modality through effective cross-modal alignment. Similarly, under landmark perturbations with $\sigma = 0.05$, UMCL consistently outperforms CPML by 33--44\% across different compression levels.
\\\indent
These results validate that UMCL's unimodal-generated multimodal approach, combined with ASA and CQSL, creates inherent redundancy and complementarity among modalities. When one modality degrades, the remaining modalities can compensate through learned cross-modal relationships, ensuring stable detection performance. This robustness makes UMCL particularly suitable for practical deployment in variable-quality media environments where data reliability fluctuates across different social media platforms and real-world scenarios.

\begin{figure*}[t!]
		\centering
		\includegraphics[width=0.95\textwidth]{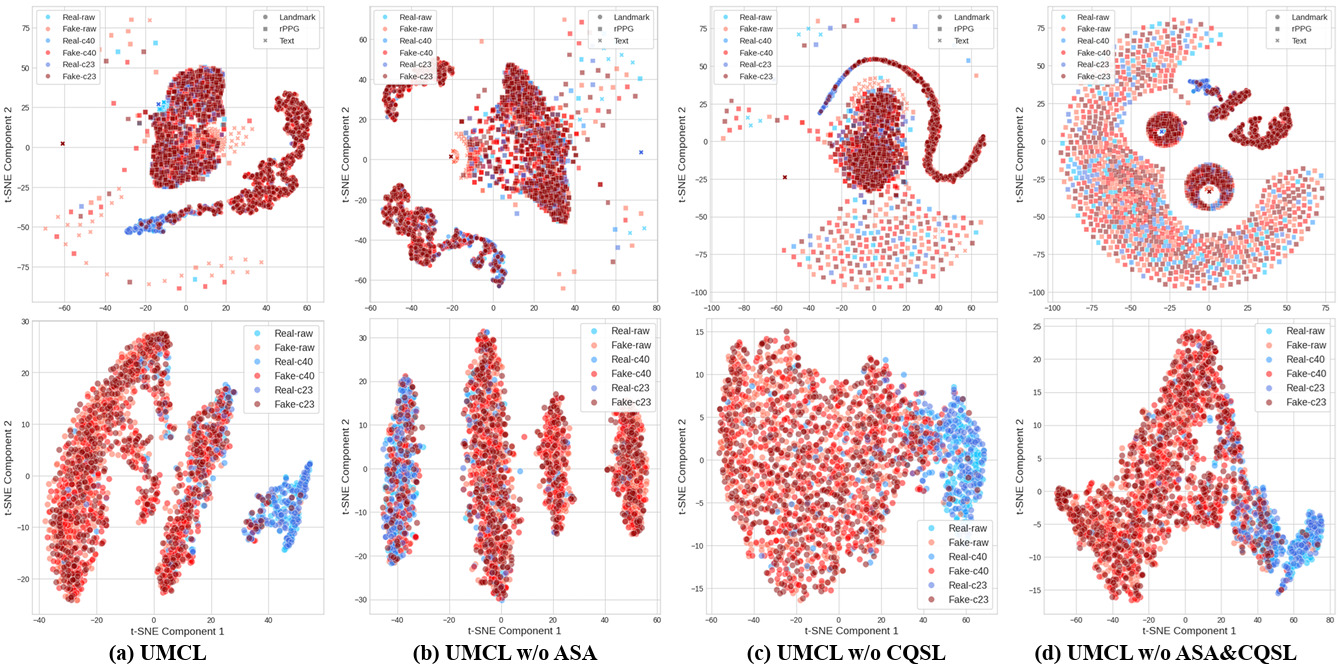} 	
        \caption{t-SNE visualizations of multimodal feature distributions under different ablation settings. The top panel shows features from different modalities before fusion. The bottom panel shows the aligned features in the fully-connected layer. 
  (a) UMCL (Full model): Features from different modalities (rPPG, landmarks, text prompts) are tightly clustered and semantically aligned, showing clear separation between real and fake samples even across compression rates.
(b) UMCL w/o ASA: Removing the ASA causes cross-modality feature drift and partial overlap between real and fake clusters, suggesting weaker consistency between modalities.
(c) UMCL w/o CQSL: Without CQSL, samples of the same content under low- and high-compression ratios fail to align properly, producing fragmented decision boundaries and modality-dependent variance.
(d) UMCL w/o ASA \& CQSL: The absence of both mechanisms results in the most scattered and indistinguishable feature space, where real/fake boundaries blur and inter-modality semantic gaps widen.
Overall, ASA strengthens cross-modality consistency, CQSL enforces cross-quality alignment, and their synergy yields compact, semantically coherent clusters that sustain discriminative separability even under degradation.}

        \label{fig:tsne}
\end{figure*}

\begin{figure*}[t]
\begin{center}
\scalebox{0.535}{
    \renewcommand{\arraystretch}{0.4}
    \begin{tabular}[c]{c@{ }c@{ }c@{ }c@{ }c@{ }c@{ }}
        \hspace*{-8mm}\includegraphics[width=0.45\textwidth,valign=t]{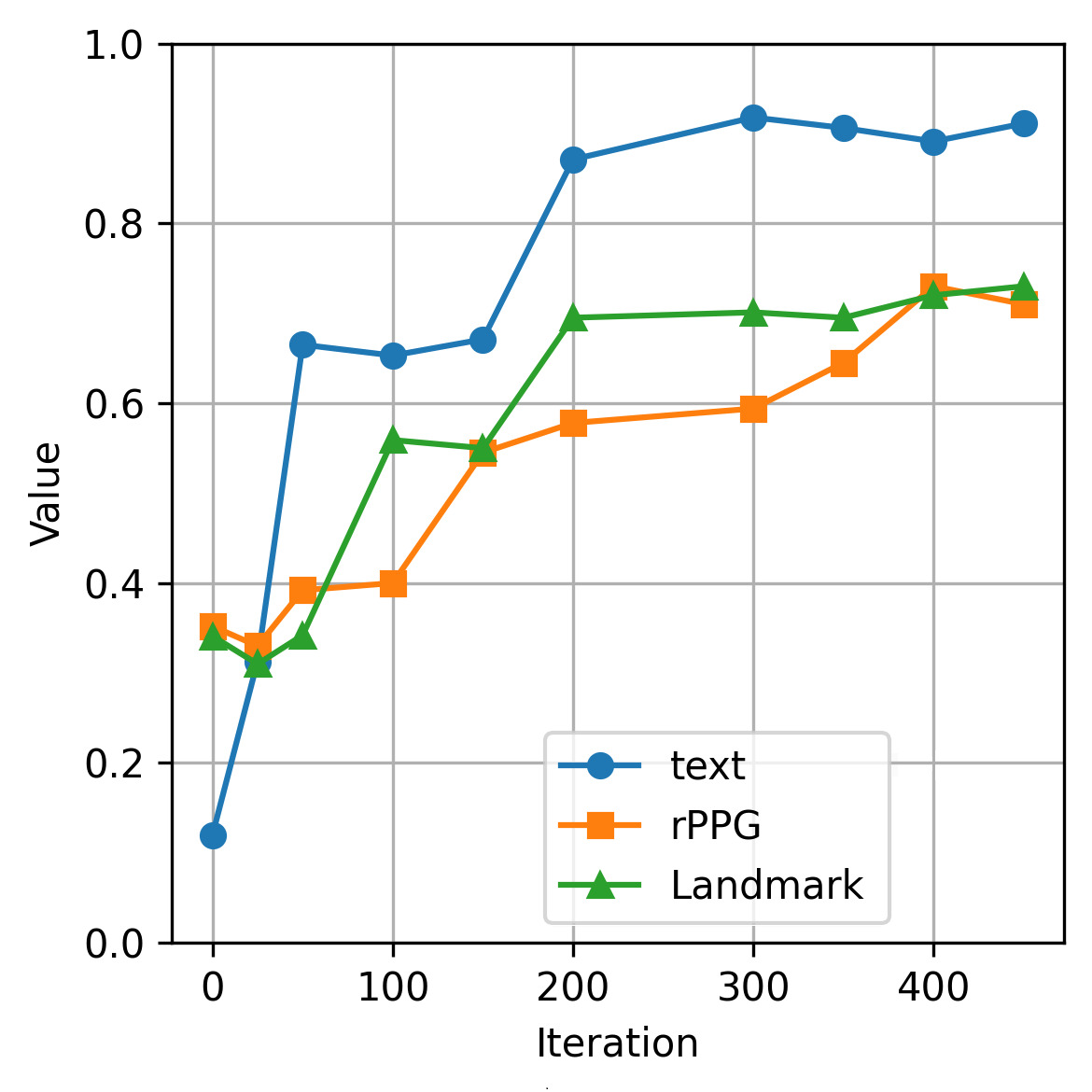}&
        \includegraphics[width=0.45\textwidth,valign=t]{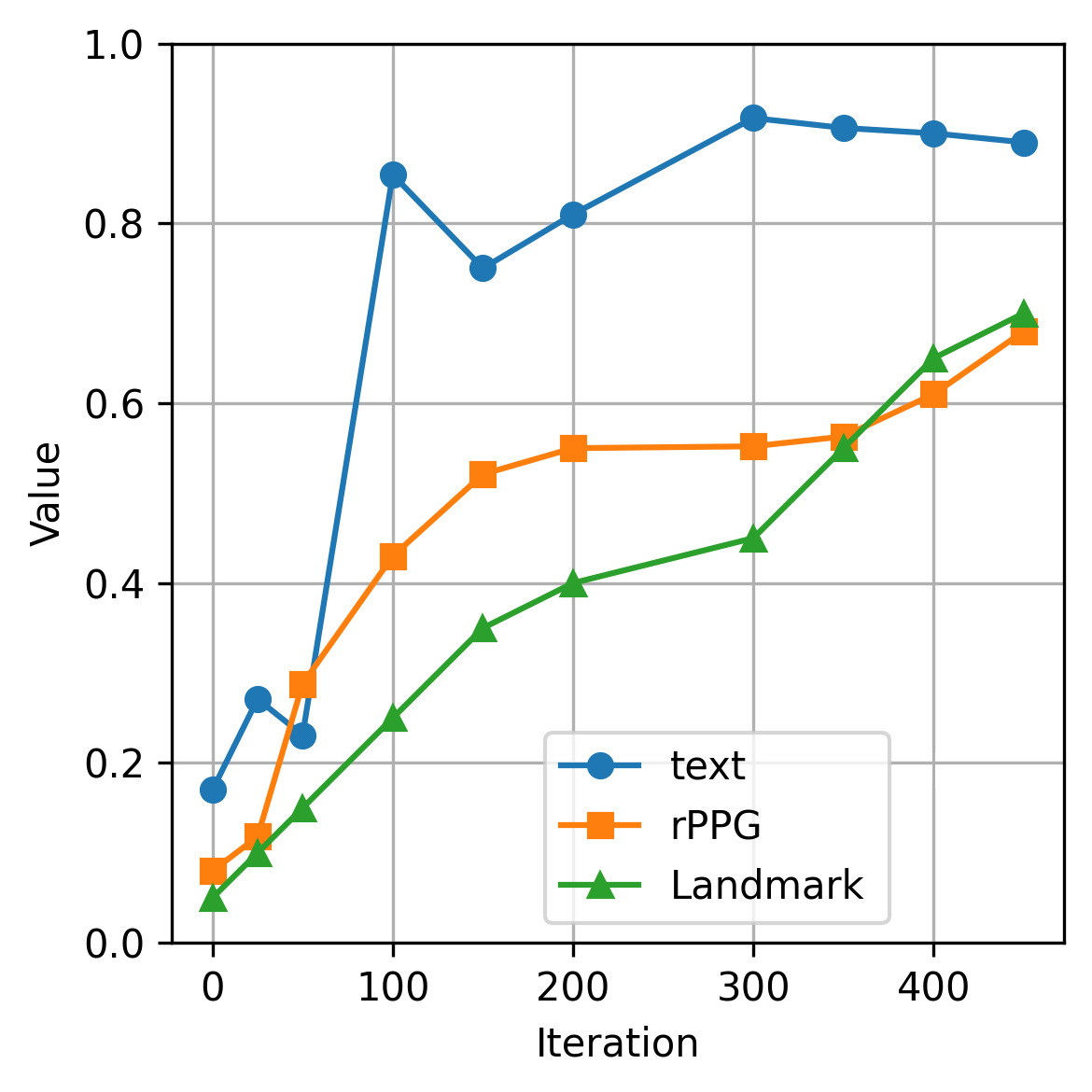}&   
        \includegraphics[width=0.45\textwidth,valign=t]{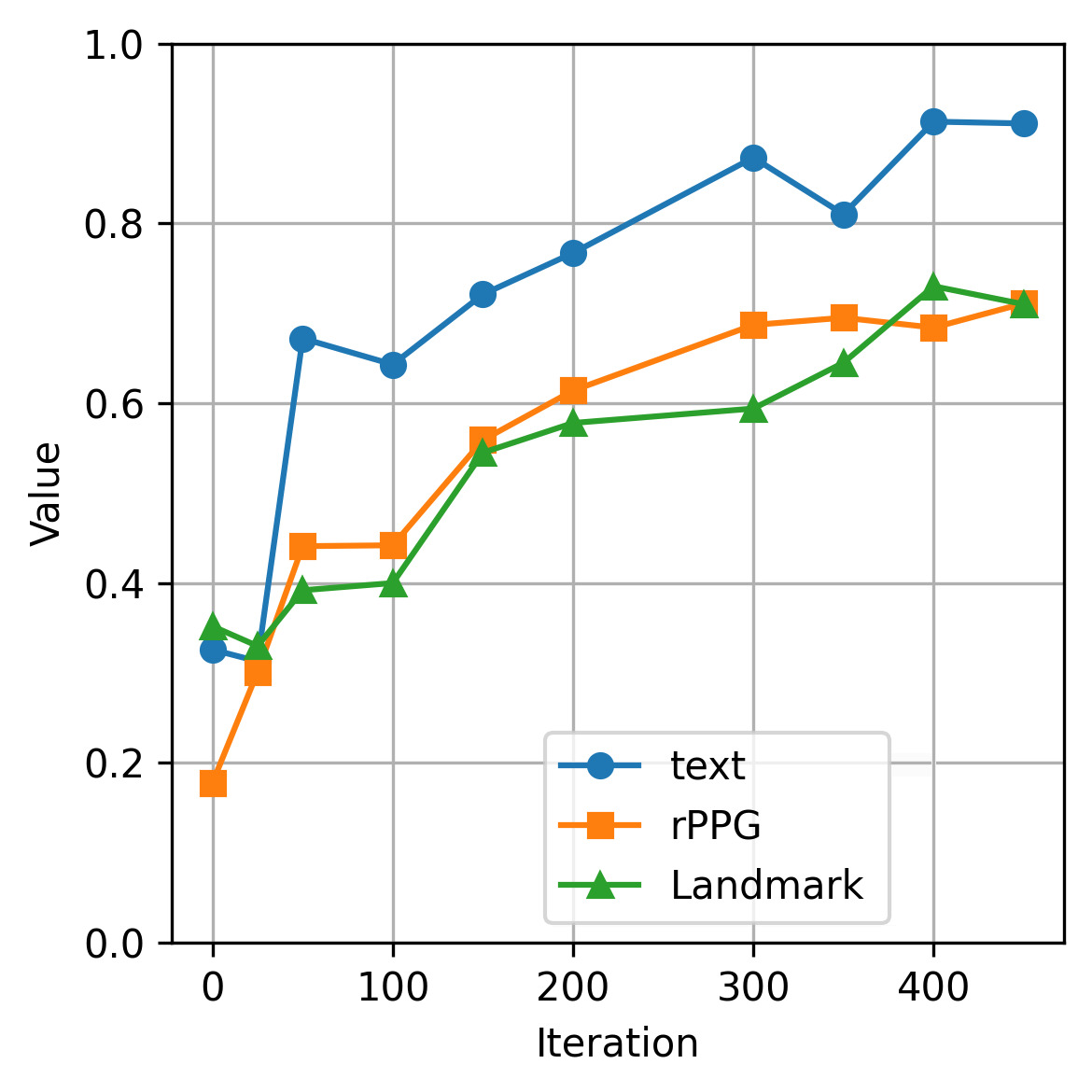}&
        \includegraphics[width=0.45\textwidth,valign=t]{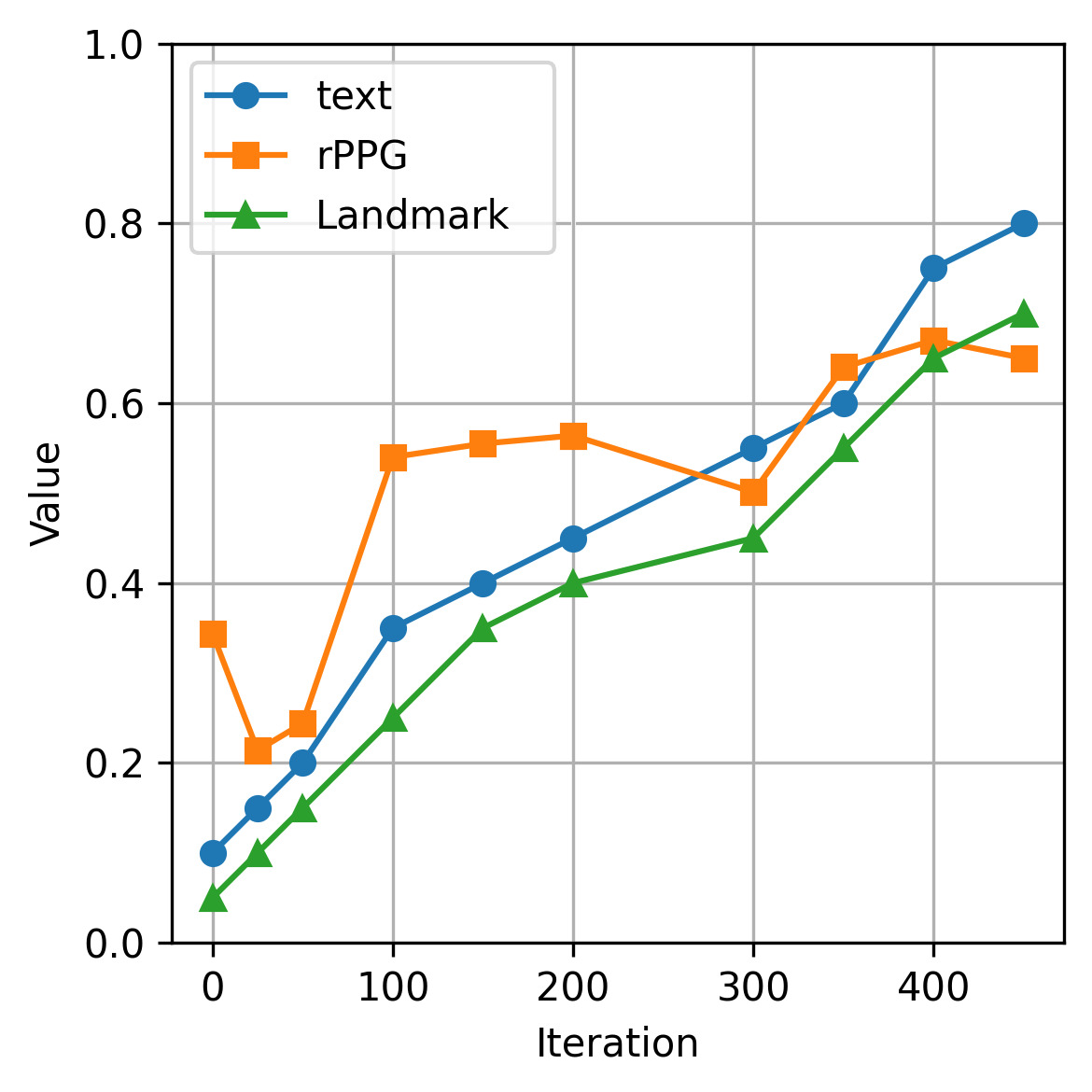}
        \\
        \hspace*{-8mm}\Large~(a) HQ-real affinity scores & \Large~(b) HQ-fake affinity scores & \Large~(c) LQ-real affinity scores & \Large~(d) LQ-fake affinity scores
    \end{tabular}
    }
\end{center}
\caption{Affinity scores between $\{\text{HQ}, \text{LQ}\}$ and $\{\text{real}, \text{fake}\}$ videos during training time.}
\label{fig:ASA_vis}
\end{figure*}

\begin{table*}[t]
\centering
\caption{Adversarial prompt robustness evaluation on FF++ across different compression levels.}
\label{tab:prompt_robustness}
\scalebox{0.85}{
\begin{tabular}{c|l|cc|cc|cc}
\hline
\multirow{2}{*}{Method} & \multirow{2}{*}{Prompt Type} & \multicolumn{2}{c|}{FF++ (raw)} & \multicolumn{2}{c|}{FF++ (c23)} & \multicolumn{2}{c}{FF++ (c40)} \\
\hhline{~~------}
& & AUC & Gain & AUC & Gain & AUC & Gain \\
\hline
\multirow{4}{*}{CPML} & \cellcolor{incompression_color}Simple Prompt& \cellcolor{incompression_color}88.46 & \cellcolor{incompression_color}- & \cellcolor{incompression_color}86.17 & \cellcolor{incompression_color}- & \cellcolor{incompression_color}65.75 & \cellcolor{incompression_color}- \\
& \cellcolor{incompression_color}Description Prompt& \cellcolor{incompression_color}89.42 & \cellcolor{incompression_color}- & \cellcolor{incompression_color}85.61 & \cellcolor{incompression_color}- & \cellcolor{incompression_color}63.03 & \cellcolor{incompression_color}- \\
& \cellcolor{incompression_color}Unrelated Prompt & \cellcolor{incompression_color}48.27 & \cellcolor{incompression_color}- & \cellcolor{incompression_color}46.18 & \cellcolor{incompression_color}- & \cellcolor{incompression_color}42.46 & \cellcolor{incompression_color}- \\
& \cellcolor{incompression_color}Opposite Prompt& \cellcolor{incompression_color}10.97 & \cellcolor{incompression_color}- & \cellcolor{incompression_color}9.82 & \cellcolor{incompression_color}- & \cellcolor{incompression_color}8.46 & \cellcolor{incompression_color}- \\
\hline
\multirow{4}{*}{UMCL} & \cellcolor{crosscompression_color}Simple Prompt& \cellcolor{crosscompression_color}96.85 & \cellcolor{crosscompression_color}{\color{deepgreen}+8.47} & \cellcolor{crosscompression_color}94.28 & \cellcolor{crosscompression_color}{\color{deepgreen}+8.11} & \cellcolor{crosscompression_color}80.54 & \cellcolor{crosscompression_color}{\color{deepgreen}+14.79} \\
& \cellcolor{crosscompression_color}Description Prompt& \cellcolor{crosscompression_color}96.85 & \cellcolor{crosscompression_color}{\color{deepgreen}+7.43} & \cellcolor{crosscompression_color}96.29 & \cellcolor{crosscompression_color}{\color{deepgreen}+10.68} & \cellcolor{crosscompression_color}80.72 & \cellcolor{crosscompression_color}{\color{deepgreen}+17.69} \\
& \cellcolor{crosscompression_color}Unrelated Prompt& \cellcolor{crosscompression_color}93.29 & \cellcolor{crosscompression_color}{\color{deepgreen}+45.02} & \cellcolor{crosscompression_color}91.05 & \cellcolor{crosscompression_color}{\color{deepgreen}+44.87} & \cellcolor{crosscompression_color}87.74 & \cellcolor{crosscompression_color}{\color{deepgreen}+45.28} \\
& \cellcolor{crosscompression_color}Opposite Prompt& \cellcolor{crosscompression_color}94.49 & \cellcolor{crosscompression_color}{\color{deepgreen}+83.52} & \cellcolor{crosscompression_color}92.42 & \cellcolor{crosscompression_color}{\color{deepgreen}+82.60} & \cellcolor{crosscompression_color}88.61 & \cellcolor{crosscompression_color}{\color{deepgreen}+80.15} \\
\hline
\end{tabular}
}
\end{table*}

\subsection{Analyses}

\textbf{t-SNE Feature Distribution.} To provide deeper insights into the effectiveness of our proposed alignment strategies, we visualize the feature distributions using the FF++ raw dataset~\cite{rossler2019faceforensics++}. Specifically, we train our model on FF++ (raw) and evaluate it across different compression levels (raw $\rightarrow$ [raw, c23, c40]), thereby offering a comprehensive view of its cross-compression generalization capability. Fig.~\ref{fig:tsne} presents t-SNE visualizations comparing the feature distributions with the ASA alignment and CQSL similarity learning strategies [Fig.~\ref{fig:tsne}(a)], UMCL without ASA [Fig.~\ref{fig:tsne}(b)], UMCL without CQSL [Fig.~\ref{fig:tsne}(c)], and UMCL without ASA and CQSL[Fig.~\ref{fig:tsne}(d)]. The features extracted by the aligned model in Fig.~\ref{fig:tsne}(a) from different modalities (rPPG, landmarks, text prompts) are tightly clustered and semantically aligned, showing clear separation between real and fake samples even across compression rates. In contrast, Fig.~\ref{fig:tsne}(b) shows that removing ASA causes cross-modality feature drift and partial overlap between real and fake clusters, suggesting weaker consistency between modalities. As shown in Fig.~\ref{fig:tsne}(c), removing CQSL prevents samples of the same content under low- and high-compression ratios from aligning properly. This misalignment results in fragmented decision boundaries and increased modality-dependent variance. Furthermore, the unaligned model in Fig.~\ref{fig:tsne}(d) exhibits a degraded feature space with substantial overlap between real and fake distributions, resulting in poor clustering quality. Overall, ASA strengthens cross-modality consistency, CQSL enforces cross-quality alignment, and their synergy yields compact, semantically coherent clusters that sustain discriminative separability even under degradation. This comparison underscores the critical roles of ASA and CQSL in learning discriminative feature representations.\\
\noindent\textbf{Affinity Score Evolution.} Fig.~\ref{fig:ASA_vis} illustrates the evolution of affinity scores between HQ/LQ and real/fake samples during training on FF++\cite{rossler2019faceforensics++}. Across all conditions, stable convergence is achieved, with HQ inputs stabilizing more quickly than LQ inputs due to their more reliable feature extraction. LQ conditions exhibit higher training variance, highlighting the need for CQSL to explicitly enforce cross-quality alignment. The affinity scores also reveal modality-specific behaviors: text embeddings demonstrate the highest stability, reflecting their semantic robustness to compression artifacts; rPPG signals display more dynamic patterns, consistent with physiological constraints and temporal dependencies; and landmark features show moderate stability, balancing structural preservation with compression resilience. Furthermore, fake samples require more training iterations to reach stable affinity alignment compared to real samples, suggesting that manipulated content introduces more complex cross-modal relationships that demand extended learning for effective alignment.
\\
\noindent\textbf{Computational Complexity.} We evaluate the computational efficiency of UMCL against the baseline CPML. As shown in Table~\ref{tab:complexity_updated}, UMCL employs a deeper architecture with 175.82M parameters compared to CPML’s 149.85M, leading to slightly higher computational demands (0.141 vs. 0.106 GFLOPs per sample) and increased memory usage (1241.5MB vs. 1142.4MB). This results in a moderate reduction in throughput from 2.95M to 2.22M samples per second. Nevertheless, UMCL’s architectural enhancements---including streamlined feature fusion and optimized affinity computations---deliver substantial accuracy improvements that justify the added overhead, making the framework significantly more robust and reliable for deepfake detection in practical deployment scenarios.
\\\indent
\begin{table}[t!]
\centering
\caption{Computational complexity comparison between CPML and UMCL (excluding T-encoder).}
\label{tab:complexity_updated}
\begin{tabular}{lcc}
\hline
\textbf{Metric} & \textbf{CPML} & \textbf{UMCL} \\
\hline
Parameters (M) & 149.85 & 175.82 \\
GFLOPs per sample & 0.106 & 0.141 \\
Throughput (samples/sec) & 2,953,211 & 2,215,304 \\
Memory Usage (FP32, MB) & 1142.4 & 1241.5 \\
\hline
\end{tabular}
\end{table}

\subsection{Ablation Studies}
\textbf{Effectiveness of Individual Modalities.}
To assess the contributions of individual modality and the alignment strategy, we conduct an ablation study on the FF++ (c23) dataset. As shown in Table~\ref{ablation_proposed}, we evaluate the following configurations:
\begin{itemize}
\item \textbf{Model A}: This model relies solely on the L-encoder to capture facial landmark dynamics. While it achieves notable performance on the DF subset, its generalization ability to other manipulation types is limited. This suggests that additional modalities are necessary to capture diverse forgery patterns.
   \item \textbf{Model B}: Similar to Model A, using only the P-encoder struggles to generalize across different manipulation subsets. This indicates that relying on a single modality, even with the CQSL strategy, is insufficient for robust deepfake detection.
   \item \textbf{Model C}: This model incorporates the outputs of L-encoder, P-encoder, and T-encoder using direct concatenation with CQSL applied, yielding significant improvement in performance.
   The average AUC increases by 4.44\% and 21.26\% compared to Model A and Model
   B, respectively.
   \item \textbf{Model D}: The full UMCL framework, incorporating the L-encoder, P-encoder, and T-encoder, achieves the best overall performance with both ASA and CQSL applied. Compared to Model C, UMCL enhances both detection accuracy and generalization, yielding an average AUC improvement of 2.99\%. This performance gain is largely attributed to the integration of textual prompts via the T-encoder and the ASA strategy, which improves feature consistency and guidance during learning.
\end{itemize}

These findings highlight the importance of leveraging multiple modalities and demonstrate the effectiveness of the proposed alignment strategies in improving deepfake detection and generalization capabilities.\\

\noindent\textbf{Effectiveness of Contrastive Learning.}
To examine the effectiveness of contrastive learning of UMCL, we compare two approaches: 
\begin{itemize}
    \item \textbf{Baseline}: Applying contrastive learning directly on image pairs corresponding to compression ratios c23 and c40
    \item \textbf{UMCL}: Employing downsampled images and landmark features with added Gaussian noise to approximate compression degradation for contrastive learning
\end{itemize}

Table~\ref{ablation_DifferentInputs} shows that UMCL performs comparably to the baseline, with only a marginal 0.85\% decrease in average AUC. This demonstrates that UMCL's contrastive learning strategy is robust to variations in compression rates while maintaining high detection performance.

More importantly, using downsampled images and landmark features as views reduces computational cost and memory usage, making UMCL highly efficient for real-world deployment. In deepfake detection systems, where models must process videos across diverse compression levels, this efficiency is critical. Despite the minor reduction in AUC (0.85\%), the significant computational savings reinforce UMCL's practicality in real-world scenarios.

\begin{table}[t!]
	\centering   	
        \caption{Ablation study comparing the effectiveness of individual modalities, and alignment strategies in the proposed UMCL, assessing performance based on AUC.}
        \begin{tabular}{|c|cccc|c|}
            \hline
        		Model & DF & F2F & FS & NT & Avg.  \\ \Xhline{1pt} 
        		A & 99.73 & 81.08 & 94.78  & 72.87 & 87.11 \\ \Xhline{1pt} 
        		B & 97.84 & 59.26 & 72.11  & 51.96 & 70.29 \\ \Xhline{1pt} 
                C & 96.82 & 91.07 & 88.93  & 89.41 & 91.55  \\\Xhline{1pt} 
                D & 96.56 & 93.27 & 97.93  & 90.41 & 94.54  \\ \hline
        	\end{tabular}
    \label{ablation_proposed}
\end{table}

\begin{table}[t]
	\centering
        \caption{Ablation study comparing the effectiveness of the different views for contrastive learning, assessing performance based on AUC (\%).}
        \begin{tabular}{|c|cccc|c|}
        	\hline
        	Model & DF & F2F & FS & NT & Avg.\\ \hline 
        	c23 \& c40 & 98.54 & 96.59 & 95.85 & 96.73  & 96.92 \\ \hline
        	Views & 97.52 & 95.88 & 95.68 & 95.22 & 96.07 \\ \hline
        	Gain & {\color{deepred}-1.02} & {\color{deepred}-0.71} & {\color{deepred}-0.17}  &{\color{deepred}-1.47}  & {\color{deepred}-0.85} \\ 
            \hline
        	\end{tabular}
    \label{ablation_DifferentInputs}
\end{table}

	\section{Conclusion}
	\label{sec:conclusion}
	This paper introduces a novel framework Unimodal-generated Multimodal Contrastive Learning (UMCL), addressing key challenges in cross-compression-rate deepfake detection. Our approach transforms a single visual modality into three complementary features: rPPG signals, facial landmark dynamics, and semantic embeddings. Through Affinity-driven Semantic Alignment and Cross-Quality Similarity Learning strategies, UMCL achieves robust alignment of these features across varying compression rates.
Extensive experiments demonstrate UMCL's superior performance in cross-compression-rate, cross-dataset, and cross-manipulation evaluations, establishing a new benchmark for deepfake detection on social media platforms. Our method maintains high detection accuracy even under severe quality degradation while providing interpretable insights through explicit feature alignment. This work represents a significant advancement toward practical, robust deepfake detection in real-world environments.

\backmatter

\section*{Acknowledgment}

This study was supported in part by the National Science and Technology Council (NSTC), Taiwan, under grants NSTC 112-2221-E-007-077-MY3, 113-2634-F-002-003, and 113-2627-M-006-005. We thank National Center for High-performance Computing (NCHC) of National Applied Research Laboratories (NARLabs) in Taiwan for providing computational and storage resources. We thank Yi-Fan Wang for her assistance in part of the experiments.

\section*{Declarations}
\begin{itemize}
\item Competing interests: The authors declare that the research was conducted in the absence of any commercial or financial relationships that could be construed as a potential conflict of interest.
\item Authors' contributions: The conceptualization and methodology were jointly developed by Ching-Yi Lai, Pei-Cheng Chuang, and Chih-Chung Hsu, with substantial contributions to material preparation, data collection, and analysis. Both Chiou-Ting Hsu and Chia-Wen Lin, as project leaders, engaged in detailed discussions regarding the project's feasibility and manuscript enhancement. Chih-Yu Jian, Chia-Ming Lee, and Yi-Fan Wang contributed to additional data analysis and validation, providing critical feedback during manuscript preparation. The first draft of the manuscript was composed by Ching-Yi Lai, and all authors contributed to the subsequent iterations through substantial comments and editing. All authors have read and agreed to the published version of the manuscript.
\item Availability of data and materials: The training data can be downloaded from \url{https://github.com/ondyari/FaceForensics} and \url{https://github.com/yuezunli/celeb-deepfakeforensics}.
\item Code availability: Codes will be made publicly available after the paper gets accepted.   
\end{itemize}

\bibliography{DeepFake}

\end{document}